\pdfoutput=1
\documentclass[10pt,twocolumn,letterpaper]{article}

\usepackage[pagenumbers]{preprint}

\usepackage{colortbl}

\newcommand{\rankfirst}[1]{\cellcolor{red!30}#1}
\newcommand{\ranksecond}[1]{\cellcolor{yellow!60}#1}

\makeatletter
\newcommand{\teasercaptype}{\def\@captype{figure}}
\makeatother
\newsavebox{\teaserimgs}

\newcommand{\p}{\mathbf{p}}
\newcommand{\uu}{\mathbf{u}}
\newcommand{\vv}{\mathbf{v}}
\newcommand{\vecop}{\operatorname{vec}}

\definecolor{linkblue}{rgb}{0.21,0.49,0.74}
\usepackage[pagebackref,breaklinks,colorlinks,allcolors=linkblue]{hyperref}
\hypersetup{pdftitle={VeilGS: Solving Darkness as a Veil of 2D Basis Maps over Retinex 3DGS},pdfauthor={Mohab Kishawy, Chengzhou Tang, Shahram Shirani, Jun Chen}}

\title{VeilGS: Solving Darkness as a Veil of 2D Basis Maps over Retinex 3DGS}

\author{%
Mohab Kishawy$^{1,\dagger}$\qquad
Chengzhou Tang$^{2}$\qquad
Shahram Shirani$^{1}$\qquad
Jun Chen$^{1}$\\[0.3em]
{\normalsize $^{1}$Department of Electrical \& Computer Engineering, McMaster University, Hamilton, Ontario, Canada}\\
{\normalsize $^{2}$Department of Computer Science, University of Manitoba, Winnipeg, Manitoba, Canada}\\[0.3em]
{\small $^{\dagger}$Corresponding author: \href{mailto:kishawym@mcmaster.ca}{\texttt{kishawym@mcmaster.ca}}}\\[0.1em]
{\small Project page: \url{https://errorlogic1211.github.io/VeilGS/}}%
}

\begin{document}

\twocolumn[{%
\renewcommand\twocolumn[1][]{#1}%
\maketitle
\vspace{-22.5pt}%
\centering
\setlength{\tabcolsep}{0.5pt}
\renewcommand{\arraystretch}{0.5}
\sbox{\teaserimgs}{\small\raisebox{\depth}{\begin{tabular}{cccc}
Input & Veil $\alpha$ (ours) & VeilGS (ours) & Gaussian-DK~\cite{Ye2024GaussianDK}\\
\includegraphics[width=0.1428\textwidth]{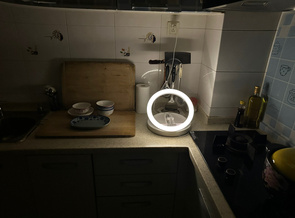} & \includegraphics[width=0.1428\textwidth]{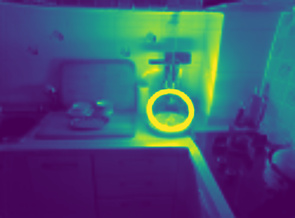} & \includegraphics[width=0.1428\textwidth]{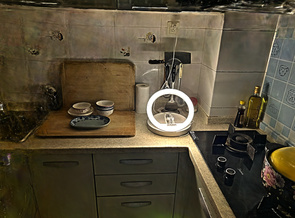} & \includegraphics[width=0.1428\textwidth]{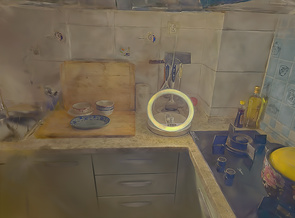}\\
\includegraphics[width=0.1428\textwidth]{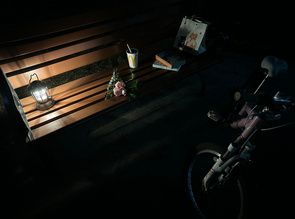} & \includegraphics[width=0.1428\textwidth]{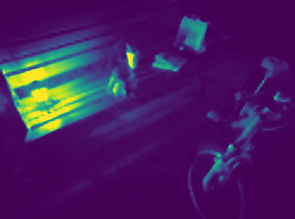} & \includegraphics[width=0.1428\textwidth]{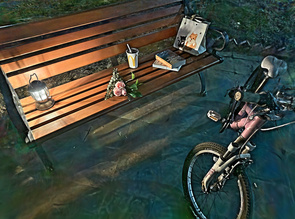} & \includegraphics[width=0.1428\textwidth]{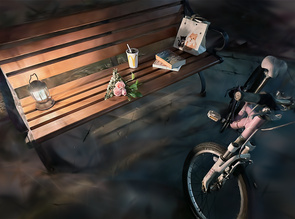}\\
\end{tabular}}}%
\begin{minipage}[b]{\wd\teaserimgs}
\centering
\usebox{\teaserimgs}\\[-3pt]
{\small (a) Visual results}
\end{minipage}\hfill
\begin{minipage}[b]{\dimexpr\textwidth-\wd\teaserimgs-6pt\relax}
\centering
\includegraphics[height=\dimexpr\ht\teaserimgs+\dp\teaserimgs\relax,width=\linewidth,keepaspectratio]{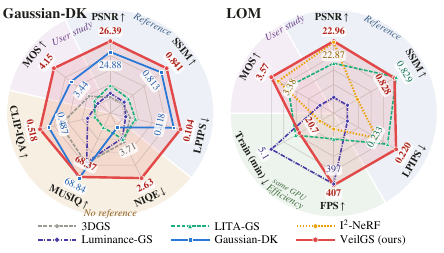}\\[-3pt]
{\small (b) Quantitative scores}
\end{minipage}
\captionsetup{hypcap=false}\teasercaptype
\caption{\textbf{VeilGS treats low-light degradation as a veil drawn over a
well-lit scene.} (a)~Held-out views of Gaussian-DK~\cite{Ye2024GaussianDK}: the solved
veil is high where light falls and low in shadow.
(b)~Mean scores, efficiency and user-study MOS; outward is better.}
\label{fig:teaser}
\vspace{1pt}%
}]

\begin{abstract}
In a dark capture, the darkness is not uniform. How much light reaches a surface depends
on where it lies relative to the few light sources, and the camera compresses what remains
with a tone curve. Most low-light 3D reconstruction methods either apply one curve to every
pixel of a view, so a dark object beside a lamp and a bright one in an unlit corner 
receive the same correction, or store the darkness in the 3D
representation, where it can absorb scene detail. We present VeilGS, a 3D Gaussian
Splatting method for low-light novel view synthesis that treats the degradation as a veil
drawn over a clear scene. Each Gaussian carries an albedo and a shading latent instead of
spherical harmonics, separating surface colour from illumination. Each view is darkened by
a spatially varying gain followed by a global tone curve, and rendering without this veil
gives the restored scene. A small network predicts a coarse basis for the gain from the
observation, and its coefficients are solved for each view
rather than learned. Training through this solve discourages the Gaussians from
absorbing what the veil can explain. VeilGS needs no paired normal-light supervision and stores no per-view parameters.
On Gaussian-DK, its restored renders score best on no-reference NIQE and CLIP-IQA
and are rated highest in a blind user study with 55 participants. On LOM, it achieves
results competitive with the state of the art.
\end{abstract}

\section{Introduction}
\label{sec:intro}

Neural Radiance Fields (NeRF)~\cite{Mildenhall2021NeRF} and 3D Gaussian Splatting
(3DGS)~\cite{Kerbl2023} have brought unprecedented progress to novel view
synthesis, with applications from virtual reality to autonomous driving and
robotics. However, both methods require well-exposed inputs. In settings such as nighttime driving,
surveillance and robotic exploration of unlit spaces, the captured images are
dark, where standard NeRF and 3DGS pipelines fail to produce usable novel views.

3D representations such as NeRF and 3DGS store the colour of each point as it appears to the camera, 
which causes the darkness to be baked into the representation itself. The darkness is also spatially 
non-uniform, since surfaces are located at different positions relative to the light sources.
Enhancing each image with a 2D low-light method~\cite{Guo2020ZeroDCE,Ma2022SCI,Cai2023Retinexformer}
before reconstruction is ineffective~\cite{Cui2024AlethNeRF,Wang2023LLNeRF,Cui2025LuminanceGS,Zhou2025LITAGS}, 
because each image is enhanced on its own, which breaks the multi-view consistency that 3D reconstruction relies 
on, creating floaters and colour drift.

To overcome this challenge, several methods handle the degradation inside the 3D
representation. LLNeRF~\cite{Wang2023LLNeRF} decomposes each 3D point into
view-independent and view-dependent components and Aleth-NeRF~\cite{Cui2024AlethNeRF}
explains darkness with concealing fields, but both inherit the slow training and
rendering of NeRF. On 3DGS, Gaussian-DK~\cite{Ye2024GaussianDK} decodes a lightness map
from per-Gaussian light features and the exposure metadata of each image,
Luminance-GS~\cite{Cui2025LuminanceGS} learns enhanced Gaussian colours from training
views corrected by a per-view colour transform and tone curve, and
LITA-GS~\cite{Zhou2025LITAGS} renders an illumination map from per-Gaussian features,
guided by illumination-invariant structure priors. These models have two limitations.
First, a curve applied uniformly over a view, as in Luminance-GS, maps each observed
intensity to the same output wherever it occurs. Unless the darkening is global, a dark
object beside a lamp and a bright object in an unlit corner may record the same value,
yet need different corrections (Fig.~\ref{fig:teaser}). Second, a spatially varying degradation stored
in the Gaussians, as in Gaussian-DK and LITA-GS, has the same resolution as the scene,
so it can become entangled with scene detail.

These limitations raise a key question: \emph{can low-light degradation be modelled as a
spatially varying, physically motivated quantity that is inferred from each
view rather than learned for it, yet coarse enough to leave the scene's texture to the
Gaussians?}

To answer it, we propose VeilGS, which treats low-light degradation as a
\emph{veil} drawn over a well-lit scene (Fig.~\ref{fig:teaser}): the
Gaussians hold the clean scene, and the veil holds the darkness. First, we replace
the spherical harmonics of each Gaussian with an albedo and a shading latent, so
that the effect of light is separable from the physical colour of the object,
following Retinex theory~\cite{Land1977Retinex}. Second, we model the veil as a
spatially varying gain followed by a global camera tone curve, and span the gain
with a coarse basis that a small network reads off the observation. Third, rather
than learning per-view coefficients for this basis, we solve for them
from each observation with a differentiable bilinear rank-one solver. Finally, we
train the scene through the solve, which discourages the Gaussians from absorbing
what the veil can explain, in the spirit of variable
projection~\cite{Golub1973VarPro}. VeilGS needs no paired normal-light
supervision, stores no per-image parameters, and renders the restored scene without
the veil.

Our contributions are summarized as follows:
\begin{itemize}
  \item We propose VeilGS, an unsupervised 3DGS framework for low-light novel
  view synthesis that models the degradation as a veil drawn over a
  well-lit scene.
  \item We replace per-Gaussian spherical harmonics with an albedo and a shading
  latent, with priors on each that encourage separating surface colour from the
  effect of light, producing a restored radiance for the veil to act on.
  \item We represent the veil as a coarse, spatially varying gain kept outside the
  Gaussians and solved from each observation, and train the scene
  through this solve, in the spirit of variable projection. To our knowledge, VeilGS is the first
  low-light 3DGS method that solves, rather than stores, its per-view degradation.
  \item Extensive experiments on the LOM and Gaussian-DK benchmarks and a user study
  with 55 participants demonstrate that VeilGS compares favourably with state-of-the-art
  NeRF- and 3DGS-based low-light methods.
\end{itemize}

\section{Related Work}
\label{sec:related}

\subsection{Novel View Synthesis under Challenging Lighting}
2D low-light methods either build on the Retinex
model~\cite{Land1977Retinex,Guo2017LIME,Wei2018RetinexNet,Cai2023Retinexformer} or
learn to enhance without paired
data~\cite{Jiang2021EnlightenGAN,Guo2020ZeroDCE,Ma2022SCI}. Applied to each view of a
capture on its own, they break the multi-view consistency that 3D reconstruction
relies on, so recent work models the lighting inside the 3D representation.

\noindent\textbf{NeRF-based methods.}
NeRF~\cite{Mildenhall2021NeRF} renders an implicit field by volume rendering and
assumes well-exposed, photometrically consistent inputs. When the lighting changes
across views, NeRF-W~\cite{MartinBrualla2021NeRFW} absorbs the change into per-view
appearance embeddings, and BilaRF~\cite{Wang2024BilaRF} into per-view bilateral
grids that mimic the camera pipeline. Both reproduce the lighting of the inputs
rather than restore it. RAW-NeRF~\cite{Mildenhall2022RawNeRF} recovers dark scenes,
but only from RAW data. For dark sRGB captures, later methods model the darkness
itself. LLNeRF~\cite{Wang2023LLNeRF} decomposes each point into view-independent and
view-dependent components, and Aleth-NeRF~\cite{Cui2024AlethNeRF} explains darkness
with concealing fields that are removed at test time. RoSe~\cite{Li2025RoSe}
estimates an illumination transition in 3D, and
I\textsuperscript{2}-NeRF~\cite{Liu2025I2NeRF} models light--media interactions with
a unified radiative equation. All of them inherit the slow training
and rendering of implicit fields.

\noindent\textbf{3DGS-based methods.}
3DGS~\cite{Kerbl2023} replaces the implicit field with explicit anisotropic
Gaussians and tile-based rasterization, reaching comparable quality in real time.
To handle appearance variance, WildGaussians~\cite{Kulhanek2024WildGaussians} brings
the per-view appearance embeddings of NeRF-W to 3DGS. Low-light 3DGS methods model
the darkness in one of two ways. Luminance-GS~\cite{Cui2025LuminanceGS} learns
enhanced Gaussian colours from training views corrected as a whole by a per-view
colour transform and tone curve, and Luminance-GS++~\cite{Cui2026LuminanceGSpp} adds
pixel-wise residual refinement. Gaussian-DK~\cite{Ye2024GaussianDK} and
LITA-GS~\cite{Zhou2025LITAGS} instead let the darkness vary across the image through
features stored in each Gaussian. Gaussian-DK decodes a lightness map from these
features and each image's exposure metadata, and LITA-GS renders an illumination
map from them, guided by illumination-invariant structure priors and a progressive
denoiser. Beyond these, LL-Gaussian~\cite{Sun2025LLGaussian} guides a
reflectance--illumination decomposition with a diffusion prior, and
RestorGS~\cite{Qiao2025RestorGS} restores several degradations in one depth-aware
framework. VeilGS takes a third route: its veil varies across the image but stays
coarse and outside the Gaussians, and is solved from each view's image rather than
learned.

\subsection{Learned Bases and Differentiable Solvers}
A common way to keep a per-image quantity simple is to write it in a small basis.
Camera responses are well described by a low-dimensional linear
basis~\cite{Grossberg2003EMoR}, and HDRNet~\cite{Gharbi2017HDRNet} infers a coarse
grid of local affine coefficients from a downsampled input. BA-Net~\cite{Tang2019BANet}
and Tang~\etal~\cite{Tang2022ZIC} go further and solve for the coefficients rather
than predict them. Solving for the linear coefficients in closed form and optimizing
only the remaining parameters is classical variable
projection~\cite{Golub1973VarPro,Ruhe1980Separable}. Gradients can pass through such a
solve by argmin differentiation~\cite{Gould2016Argmin}, optimization
layers~\cite{Amos2017OptNet} or closed-form ridge regression~\cite{Bertinetto2019R2D2},
so the networks around it can be trained end to end. These solvers are, however,
typically trained with ground-truth supervision. VeilGS places such a solve inside the
optimization of a 3D scene, where neither the scene nor its degradation has ground truth.

\section{Method}
\label{sec:method}

\begin{figure*}[t]
\centering
\includegraphics[width=0.95\textwidth]{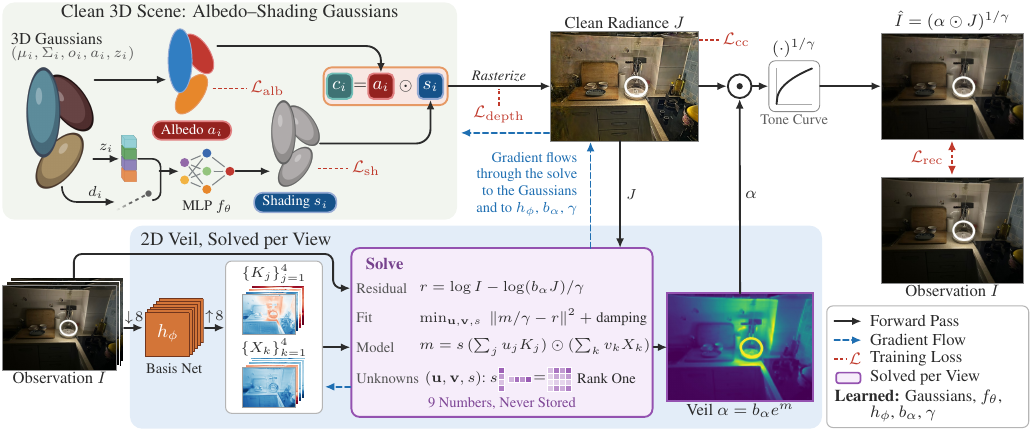}
\caption{\textbf{Overview of VeilGS.} The Gaussians render a clean radiance $J$, the
restored view.
A basis network reads the observation $I$ and proposes smooth basis maps, and a
differentiable solve finds the coefficients under which the low-light model applied to
$J$ best explains $I$, giving the gain $\alpha$ and the prediction $\hat{I}$. The
reconstruction loss is back-propagated through the solve. No clean image is used.}
\label{fig:pipeline}
\end{figure*}

Given $N$ posed low-light sRGB images $\{I_n\}_{n=1}^{N}$ of a static scene without normal-light reference, VeilGS recovers 3D Gaussians whose render $J$ is the
scene's clean radiance (Fig.~\ref{fig:pipeline}). For a training view $n$, we
rasterize $J_n$ from an albedo--shading decomposition (Sec.~\ref{sec:decomp}),
and a basis network reads $I_n$ to propose a subspace for the spatially varying
gain of the low-light forward model (Sec.~\ref{sec:veil}). The coefficients in that subspace are solved
to explain $I_n$ (Sec.~\ref{sec:solve}), the reconstruction loss is
back-propagated through the solve (Sec.~\ref{sec:varpro}), and weak global
priors select what the observations leave open (Sec.~\ref{sec:losses}).

\subsection{Preliminaries: 3D Gaussian Splatting}
\label{sec:prelim}

3DGS~\cite{Kerbl2023} represents a scene explicitly as a set of anisotropic 3D
Gaussians $\{G_i\}_{i=1}^{M}$. Each Gaussian is defined by a centre
$\mu_i\in\mathbb{R}^3$ and a covariance $\Sigma_i\in\mathbb{R}^{3\times3}$,
\begin{equation}
  G_i(\mathbf{x}) = e^{-\frac{1}{2}(\mathbf{x}-\mu_i)^{\top}\Sigma_i^{-1}(\mathbf{x}-\mu_i)},
  \label{eq:gaussian}
\end{equation}
where $\Sigma_i$ is factored into scaling $S_i$ and rotation $R_i$,
\begin{equation}
  \Sigma_i=R_iS_iS_i^{\top}R_i^{\top},
  \label{eq:covariance}
\end{equation}
which, with positive scales, keeps $\Sigma_i$ positive definite.
Each Gaussian also carries an opacity
$o_i\in[0,1]$ and a view-dependent colour $c_i$, represented by spherical
harmonics coefficients. To render a view, the Gaussians are projected onto the
image plane, and the colour of a pixel
$\p$ is obtained by alpha-blending the depth-sorted Gaussians that overlap it,
\begin{equation}
  C(\p)=\sum_{i} c_i\,\omega_i(\p)\prod_{i'<i}\big(1-\omega_{i'}(\p)\big),
  \label{eq:blend}
\end{equation}
where $\omega_i(\p)$ is the opacity $o_i$ weighted by the projected 2D Gaussian
at $\p$. The Gaussians are optimized with an $\ell_1$ and D-SSIM loss, with adaptive
densification and pruning~\cite{Kerbl2023}. VeilGS keeps this geometry,
rasterization and densification unchanged, and changes only what $c_i$ represents
(Sec.~\ref{sec:decomp}) and how the rendered radiance is transformed into the observed
low-light image
(Secs.~\ref{sec:veil}--\ref{sec:varpro}).

\subsection{Albedo--Shading Decomposition}
\label{sec:decomp}

The spherical harmonics
colour $c_i$ of Eq.~\eqref{eq:blend} describes how a
point \emph{looks}, which under a dark capture includes the darkness. We factorize it into an albedo $a_i\in\mathbb{R}^3$ and a shading
$s_i\in\mathbb{R}^3_{>0}$ produced by a small MLP $f_{\theta}$ shared by all
Gaussians,
\begin{equation}
  s_i=f_{\theta}(z_i,d_i),\qquad c_i=a_i\odot s_i,
  \label{eq:decomp}
\end{equation}
where $z_i\in\mathbb{R}^8$ is a per-Gaussian shading latent and $d_i$ is the viewing ray direction from the camera centre to the Gaussian centre $\mu_i$. Following Retinex
theory~\cite{Land1977Retinex}, the albedo carries the surface colour, while the
shading models the light it receives in the scene. Since the decomposition of $a_i\odot s_i$ is ambiguous, we regularize the decomposition with the priors introduced in Sec.~\ref{sec:losses}.
The decomposition matters because a spherical-harmonic colour can render a shadowed
surface as a dark one, whereas here the albedo should be a reflectance in $[0,1]$
similar to its neighbours and the shading should be grey and bounded, so the
shadows in the scene are separable from the surface colour.
A Softplus output makes shading positive, so it can scale albedo
but never invert it. View-dependent effects remain representable, but they now
live in the shading.

The composited colour of Eq.~\eqref{eq:blend} is the clean radiance
$J\in\mathbb{R}_{\ge 0}^{H\times W\times 3}$, a canonical restored appearance that keeps
the shading but not the per-view attenuation.

\subsection{The Veil}
\label{sec:veil}

We model a low-light image with a simple camera model, in which a pixel is the
camera's tone curve applied to the light reaching the
sensor~\cite{Debevec1997HDR,Grossberg2003EMoR}, and let that light be the clean
radiance scaled by an illumination map, as in
Retinex~\cite{Land1977Retinex,Guo2017LIME}:
\begin{equation}
  \hat{I}=\big(\alpha\odot J\big)^{\frac{1}{\gamma}},
  \label{eq:lowlight}
\end{equation}
where $\alpha\in(0,1]$ is the gain, which we call the \emph{veil}, and $1/\gamma$ is
the exponent of the camera's tone curve~\cite{Lv2021AGLLNet}, one number per scene.
In all experiments, we average this curve with a luminance-preserving one
(Sec.~\ref{sup:formation}). The veil has one value per pixel, shared by the three colour channels, so before
clipping it changes brightness but not colour. It varies across the image, since
pixels near a lamp receive more light. The restored view is $J$ itself, without the
veil and the tone curve, clipped to $[0,1]$. Because scaling $J$ up and $\alpha$ down leaves $\hat{I}$
unchanged, an exposure prior selects the brightness of $J$ (Sec.~\ref{sec:losses}).

\noindent\textbf{Representing the veil.}
The gain is a global level times a local modulation,
$\alpha=b_{\alpha}\exp(m)$, where $b_{\alpha}>0$ is learned and $m$
raises or lowers the level locally; the forward model clips $\alpha$ at 1.
The map $m$ should be smooth, which discourages it from absorbing scene texture. A small network $h_{\phi}$ reads the dark image at
one-eighth resolution and outputs two sets of $d=4$ coarse maps,
\begin{equation}
  \big\{K_j\big\}_{j=1}^{d},\;\big\{X_k\big\}_{k=1}^{d}
  =\tanh\!\Big(\mathcal{U}_{8}\,h_{\phi}\big(\mathcal{P}_{8}\,I_n\big)\Big),
  \label{eq:basis}
\end{equation}
where $\mathcal{P}_8$ and $\mathcal{U}_8$ shrink and enlarge by a factor of 8.
Following Tang~\etal~\cite{Tang2022ZIC}, $m$ combines the two sets bilinearly, with a
pixel-wise product where they use a convolution,
\begin{equation}
  m=s\Big(\sum_{j}u_jK_j\Big)\odot\Big(\sum_{k}v_kX_k\Big)=\sum_{j,k}M_{jk}\,\Phi_{jk},
  \label{eq:bilinear}
\end{equation}
with unit vectors $\uu,\vv\in\mathbb{R}^{d}$, a scale $s\ge 0$,
$\Phi_{jk}=K_j\odot X_k$ and $M=s\,\uu\vv^{\top}$. Each view's veil is thus nine
numbers $(\uu,\vv,s)$, computed at every step (Sec.~\ref{sec:solve}) rather than
stored. The reason behind choosing a rank-one product over a sum of the eight maps is
that with a similar number of free coefficients per view (seven against eight), it
draws on the 16 maps $\Phi_{jk}$ and lets one set scale the other. So the gain can
take more shapes, while the rank-one constraint on $M$ keeps it coarse, which
discourages copying scene texture.

\subsection{Solving for the Coefficients}
\label{sec:solve}

For each view, $M=s\,\uu\vv^{\top}$ is chosen so that the prediction $\hat{I}$ of
Eq.~\eqref{eq:lowlight} matches the observation $I$, with the clean render $J$, the
global level $b_{\alpha}$, the tone-curve parameter $\gamma$ and the basis maps held
fixed.

\noindent\textbf{The residual.}
Taking logs turns the multiplication by the veil into an addition. With
$\alpha=b_{\alpha}\exp(m)$, the log of Eq.~\eqref{eq:lowlight} at pixel $\p$ and
colour channel $\ell$ is
\begin{equation}
  \log\hat{I}(\p,\ell)=
  \underbrace{\tfrac{1}{\gamma}\log\big(b_{\alpha}J(\p,\ell)\big)}_{\text{known}}
  +\underbrace{\tfrac{1}{\gamma}\,m(\p)}_{\text{veil}}.
  \label{eq:loglin}
\end{equation}
Matching it to the observed $\log I(\p,\ell)$, the veil's part must equal the
\emph{residual}
\begin{equation}
  r(\p,\ell)=\log I(\p,\ell)-\tfrac{1}{\gamma}\log\big(b_{\alpha}J(\p,\ell)\big),
  \label{eq:residual}
\end{equation}
the part of the observed log image that the clean render and the global level do
not explain.

\noindent\textbf{The fit.}
Since $m$ is a weighted sum of the 16 maps, the nine numbers of the veil are found
by a rank-constrained least-squares fit,
\begin{equation}
\begin{aligned}
  \min_{s,\uu,\vv}\;&\sum_{\p,\ell}\Big(\tfrac{1}{\gamma}m(\p)-r(\p,\ell)\Big)^{2}\\
  &+\vecop(M)^{\top}\Lambda\,\vecop(M),
\end{aligned}
  \label{eq:ls}
\end{equation}
where $m$ is the bilinear form of Eq.~\eqref{eq:bilinear}, so the fit is over the
nine numbers rather than over the 16 entries of $M=s\,\uu\vv^{\top}$. If the maps
$\Phi_{jk}/\gamma$ form the 16 columns of a matrix $A$, with one row per pixel and
channel, the first term is $\|A\,\vecop(M)-r\|^2$. The second term is a Marquardt-style
penalty~\cite{Marquardt1963}. It keeps the coefficients small, penalizing each entry
of $M$ in proportion to how strongly its map acts on the image, so the fit stays well
conditioned as $\gamma$ changes.

\noindent\textbf{Solver.}
We solve Eq.~\eqref{eq:ls} in three steps, adapting the solver of
Tang~\etal~\cite{Tang2022ZIC}. (1)~Ignoring the rank constraint leaves a damped
least-squares problem, which we solve in closed form with a QR factorization.
(2)~Power iteration approximates the rank-one matrix closest to that solution, which
gives $\uu$, $\vv$ and $s$; it starts from a fixed vector rather than a random one. (3)~Four Gauss--Newton steps adjust $\uu$, $\vv$
and $s$ to fit Eq.~\eqref{eq:ls} better. The three steps run on $64\times64$ downsized
copies of $I$, $J$ and the basis maps. As the veil is shared by the channels, only the
channel mean of $r$ matters, giving $4{,}096$ equations for 16 unknowns and a few small
linear solves per training iteration. The resulting $(\uu,\vv,s)$ define the full-resolution veil
through Eq.~\eqref{eq:bilinear}. Sec.~\ref{sup:solver} details all three steps.

\subsection{Training Through the Solve}
\label{sec:varpro}

The solved coefficients depend on what we train: the Gaussians and the global level
enter the residual $r$, and the basis network and $\gamma$ enter $A$. Rather than
detaching the solve, we back-propagate through it, so the
gradient also includes how the coefficients change when the scene changes. A change
the veil can undo is then partly cancelled by the
re-fit, so the Gaussians receive less gradient to absorb it.

Without the rank constraint, the fit leaves the residual
\begin{equation}
  r^{\star}=\Big(\mathbf{I}-A\big(A^{\top}A+\Lambda\big)^{-1}A^{\top}\Big)\,r,
  \label{eq:projector}
\end{equation}
which, without damping, is exactly the part of $r$ that the veil cannot explain. In
this idealized setting, solving for the coefficients and training the rest on
$r^{\star}$ is variable projection~\cite{Golub1973VarPro}. Our fit is damped and
rank-one, and the reconstruction loss differs from its objective, so this view only
motivates training through the solve: Table~\ref{tab:abl_mech} shows its benefit
empirically, and Sec.~\ref{sup:varpro} analyses the gradient.

\noindent\textbf{Keeping the biases valid.}
Gradients through the solve can push $b_{\alpha}$ or $\gamma$ out of their valid
range. Once outside, the clipping in the forward model passes no gradient, so they
never return and the veil stops training. After every optimizer step we therefore
clip both back into range.

\begin{figure*}[t]
\centering
\setlength{\tabcolsep}{1pt}
\renewcommand{\arraystretch}{0.5}
\begin{tabular}{cccccc}
\includegraphics[width=0.162\textwidth]{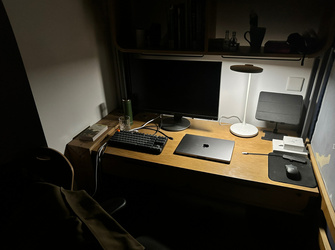} & \includegraphics[width=0.162\textwidth]{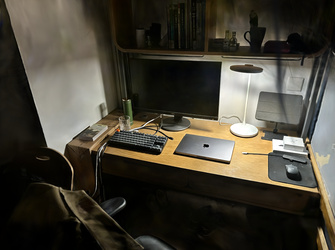} & \includegraphics[width=0.162\textwidth]{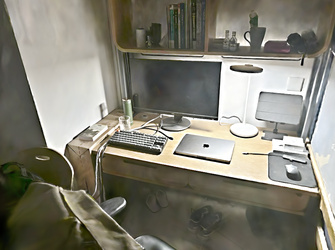} & \includegraphics[width=0.162\textwidth]{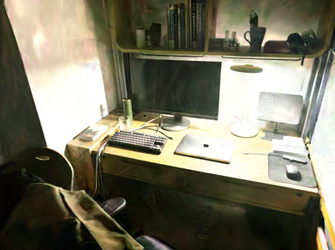} & \includegraphics[width=0.162\textwidth]{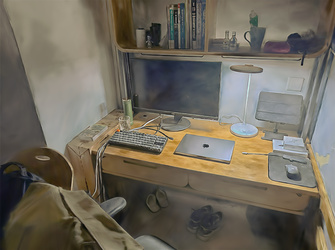} & \includegraphics[width=0.162\textwidth]{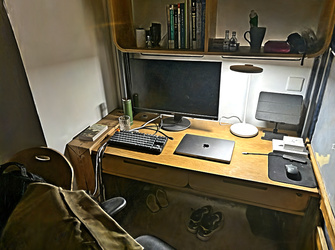} \\
\includegraphics[width=0.162\textwidth]{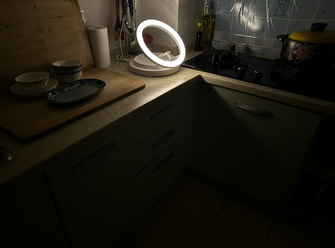} & \includegraphics[width=0.162\textwidth]{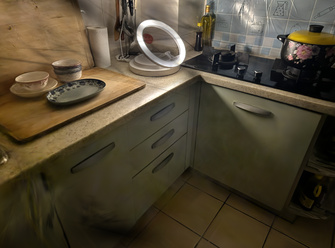} & \includegraphics[width=0.162\textwidth]{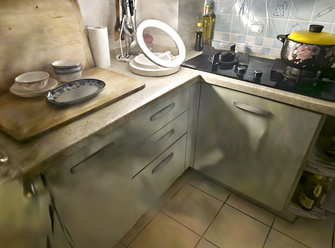} & \includegraphics[width=0.162\textwidth]{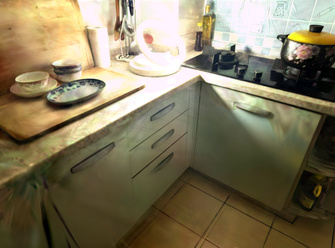} & \includegraphics[width=0.162\textwidth]{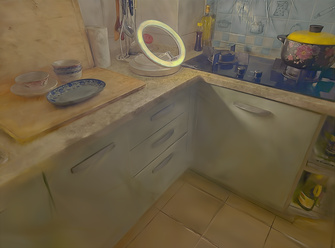} & \includegraphics[width=0.162\textwidth]{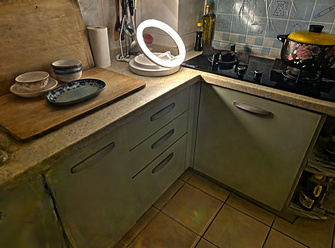} \\
\includegraphics[width=0.162\textwidth]{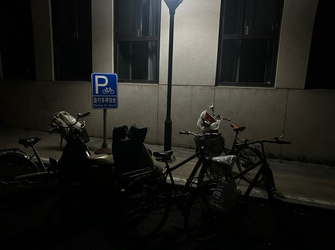} & \includegraphics[width=0.162\textwidth]{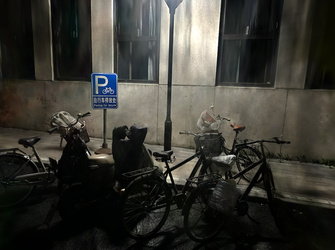} & \includegraphics[width=0.162\textwidth]{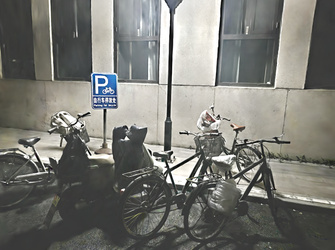} & \includegraphics[width=0.162\textwidth]{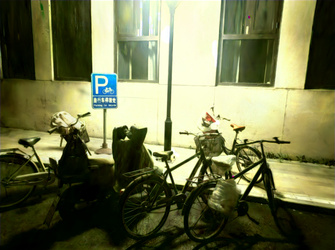} & \includegraphics[width=0.162\textwidth]{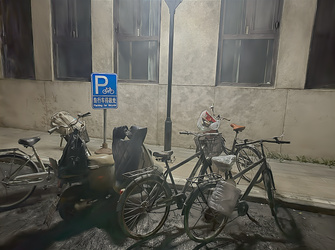} & \includegraphics[width=0.162\textwidth]{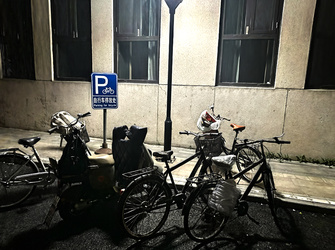} \\
\includegraphics[width=0.162\textwidth]{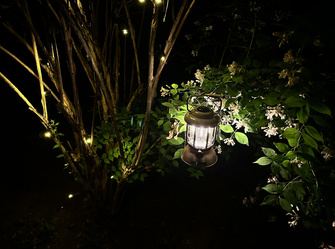} & \includegraphics[width=0.162\textwidth]{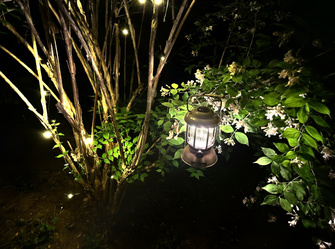} & \includegraphics[width=0.162\textwidth]{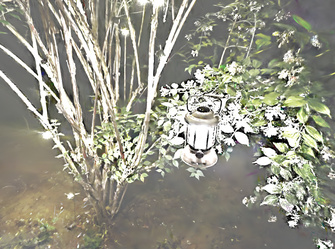} & \includegraphics[width=0.162\textwidth]{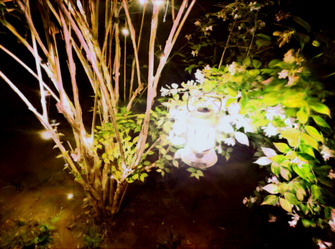} & \includegraphics[width=0.162\textwidth]{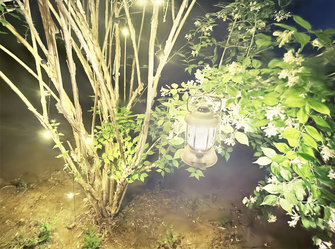} & \includegraphics[width=0.162\textwidth]{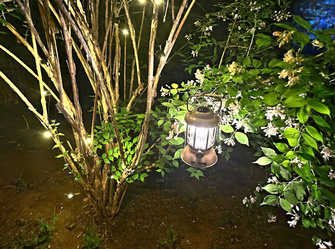} \\
\small Dark input & \small 3DGS~\cite{Kerbl2023} & \small Luminance-GS~\cite{Cui2025LuminanceGS} &
\small LITA-GS~\cite{Zhou2025LITAGS} & \small Gaussian-DK~\cite{Ye2024GaussianDK} & \small VeilGS (ours) \\
\end{tabular}
\caption{\textbf{Qualitative comparison on Gaussian-DK}: held-out views of
\textsc{dormitory}, \textsc{kitchen}, \textsc{street} and \textsc{tree}, top to bottom.}
\label{fig:qual_dk}
\end{figure*}

\begin{table*}[t]
\centering\scriptsize
\setlength{\tabcolsep}{1.2pt}
\renewcommand{\arraystretch}{1.0}
\caption{\textbf{Restoration quality on Gaussian-DK}: per-scene no-reference scores and
user-study MOS of the restored render. Red: best; yellow: second best.}
\label{tab:dk_noref}
\resizebox{\textwidth}{!}{\begin{tabular}{l|cccc|cccc|cccc|cccc|cccc}
\hline
Scene & \multicolumn{4}{c|}{3DGS~\cite{Kerbl2023}} & \multicolumn{4}{c|}{LITA-GS~\cite{Zhou2025LITAGS}} & \multicolumn{4}{c|}{Luminance-GS~\cite{Cui2025LuminanceGS}} & \multicolumn{4}{c|}{Gaussian-DK~\cite{Ye2024GaussianDK}} & \multicolumn{4}{c}{VeilGS (ours)} \\
 & {\tiny NIQE$\downarrow$} & {\tiny MUSIQ$\uparrow$} & {\tiny CLIP-IQA$\uparrow$} & {\tiny MOS$\uparrow$} & {\tiny NIQE$\downarrow$} & {\tiny MUSIQ$\uparrow$} & {\tiny CLIP-IQA$\uparrow$} & {\tiny MOS$\uparrow$} & {\tiny NIQE$\downarrow$} & {\tiny MUSIQ$\uparrow$} & {\tiny CLIP-IQA$\uparrow$} & {\tiny MOS$\uparrow$} & {\tiny NIQE$\downarrow$} & {\tiny MUSIQ$\uparrow$} & {\tiny CLIP-IQA$\uparrow$} & {\tiny MOS$\uparrow$} & {\tiny NIQE$\downarrow$} & {\tiny MUSIQ$\uparrow$} & {\tiny CLIP-IQA$\uparrow$} & {\tiny MOS$\uparrow$} \\
\hline
\textsc{alley} & 2.98 & 64.73 & \ranksecond{0.384} & 2.47 & 2.99 & 52.25 & 0.227 & 2.65 & \ranksecond{2.96} & \ranksecond{65.56} & 0.274 & 2.80 & 3.06 & \ranksecond{65.56} & 0.357 & \ranksecond{3.60} & \rankfirst{2.21} & \rankfirst{67.22} & \rankfirst{0.442} & \rankfirst{4.65} \\
\textsc{bicycle} & \ranksecond{4.27} & 61.82 & 0.423 & 2.44 & 4.42 & 45.68 & 0.280 & 2.58 & 4.43 & 56.77 & 0.230 & 1.95 & 4.91 & \ranksecond{69.29} & \ranksecond{0.472} & \ranksecond{3.84} & \rankfirst{2.94} & \rankfirst{69.64} & \rankfirst{0.492} & \rankfirst{3.95} \\
\textsc{dormitory} & 4.44 & 69.03 & \ranksecond{0.450} & 2.84 & \ranksecond{4.10} & 50.55 & 0.329 & 1.95 & 4.58 & 69.66 & 0.372 & 2.62 & 4.95 & \rankfirst{73.56} & 0.439 & \ranksecond{3.29} & \rankfirst{3.01} & \ranksecond{72.58} & \rankfirst{0.525} & \rankfirst{4.45} \\
\textsc{fence} & \ranksecond{2.93} & 55.45 & 0.494 & 2.31 & 3.54 & 45.41 & 0.376 & 2.07 & 3.18 & 59.58 & 0.436 & 2.33 & 3.39 & \rankfirst{65.47} & \ranksecond{0.541} & \ranksecond{3.53} & \rankfirst{2.34} & \ranksecond{64.77} & \rankfirst{0.555} & \rankfirst{4.24} \\
\textsc{flower} & \ranksecond{2.67} & \ranksecond{72.20} & \ranksecond{0.583} & \ranksecond{3.36} & 3.49 & 58.25 & 0.436 & 2.71 & 3.21 & 70.00 & 0.504 & 2.45 & 3.82 & \rankfirst{74.31} & \ranksecond{0.583} & 2.24 & \rankfirst{2.48} & 69.25 & \rankfirst{0.600} & \rankfirst{4.13} \\
\textsc{kitchen} & \ranksecond{3.90} & 70.30 & 0.422 & \ranksecond{3.20} & 3.94 & 53.23 & 0.292 & 2.47 & 4.20 & 69.34 & 0.389 & 2.69 & 5.18 & \rankfirst{73.86} & \ranksecond{0.454} & 2.40 & \rankfirst{2.58} & \ranksecond{72.37} & \rankfirst{0.521} & \rankfirst{4.78} \\
\textsc{livingroom} & \ranksecond{3.32} & 65.92 & \ranksecond{0.403} & 3.07 & 3.51 & 52.79 & 0.294 & 2.96 & 3.51 & 63.01 & 0.318 & \ranksecond{3.31} & 3.56 & \rankfirst{69.76} & 0.368 & 3.18 & \rankfirst{2.65} & \ranksecond{69.14} & \rankfirst{0.473} & \rankfirst{4.60} \\
\textsc{piano} & 4.73 & 52.72 & 0.441 & 1.82 & \ranksecond{4.65} & 42.54 & 0.299 & 2.24 & 5.11 & 49.11 & 0.278 & 1.82 & 5.44 & \ranksecond{62.18} & \ranksecond{0.528} & \rankfirst{4.13} & \rankfirst{2.64} & \rankfirst{65.69} & \rankfirst{0.542} & \ranksecond{3.51} \\
\textsc{snowman} & \ranksecond{4.19} & 57.17 & 0.417 & \ranksecond{3.16} & 4.44 & 46.62 & 0.371 & 2.00 & 4.28 & \ranksecond{62.79} & 0.402 & 1.98 & 4.66 & 62.03 & \ranksecond{0.447} & \rankfirst{4.45} & \rankfirst{3.13} & \rankfirst{64.12} & \rankfirst{0.464} & 2.98 \\
\textsc{staircase} & 5.41 & 58.56 & 0.384 & 2.20 & \ranksecond{3.85} & 39.37 & 0.307 & 2.29 & 4.21 & 60.65 & 0.279 & 1.80 & 4.74 & \ranksecond{69.51} & \ranksecond{0.526} & \ranksecond{3.36} & \rankfirst{2.92} & \rankfirst{69.55} & \rankfirst{0.546} & \rankfirst{3.95} \\
\textsc{street} & \ranksecond{2.48} & 64.76 & 0.411 & 2.33 & 3.30 & 48.08 & 0.284 & 2.96 & 2.78 & 61.01 & 0.334 & 2.11 & 3.06 & \rankfirst{72.19} & \rankfirst{0.541} & \ranksecond{3.73} & \rankfirst{1.95} & \ranksecond{66.71} & \ranksecond{0.491} & \rankfirst{4.29} \\
\textsc{tree} & 3.24 & 64.96 & 0.523 & 2.98 & 3.39 & 55.35 & 0.365 & 2.51 & 3.44 & 64.02 & 0.363 & 1.87 & \ranksecond{3.05} & \ranksecond{68.40} & \rankfirst{0.585} & \ranksecond{3.56} & \rankfirst{2.69} & \rankfirst{69.35} & \ranksecond{0.565} & \rankfirst{4.25} \\
\hline
Mean & \ranksecond{3.71} & 63.14 & 0.445 & 2.68 & 3.80 & 49.18 & 0.322 & 2.45 & 3.83 & 62.62 & 0.348 & 2.31 & 4.15 & \rankfirst{68.84} & \ranksecond{0.487} & \ranksecond{3.44} & \rankfirst{2.63} & \ranksecond{68.37} & \rankfirst{0.518} & \rankfirst{4.15} \\
\hline
\end{tabular}
}
\end{table*}

\begin{figure*}[t]
\centering
\setlength{\tabcolsep}{1pt}
\renewcommand{\arraystretch}{0.5}
\begin{tabular}{cccccc}
\includegraphics[width=0.162\textwidth]{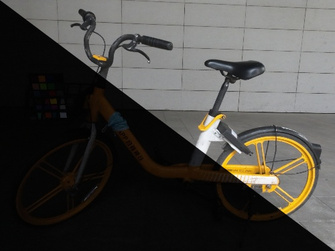} & \includegraphics[width=0.162\textwidth]{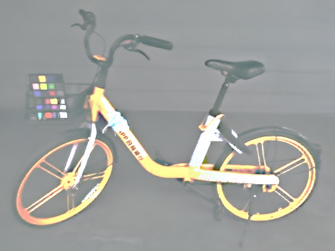} & \includegraphics[width=0.162\textwidth]{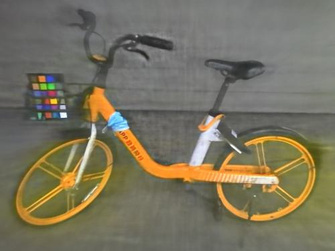} & \includegraphics[width=0.162\textwidth]{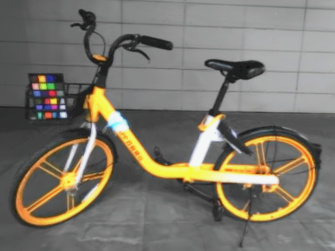} & \includegraphics[width=0.162\textwidth]{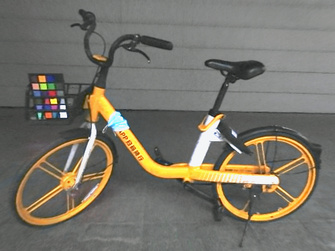} & \includegraphics[width=0.162\textwidth]{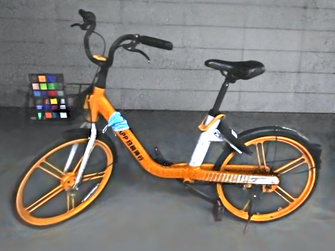} \\
\includegraphics[width=0.162\textwidth]{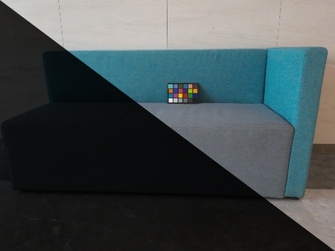} & \includegraphics[width=0.162\textwidth]{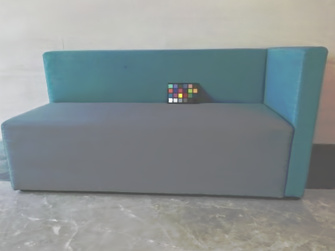} & \includegraphics[width=0.162\textwidth]{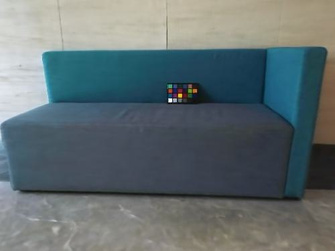} & \includegraphics[width=0.162\textwidth]{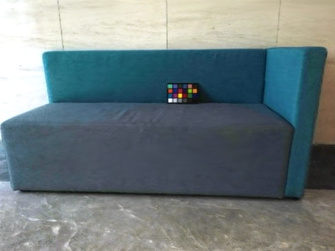} & \includegraphics[width=0.162\textwidth]{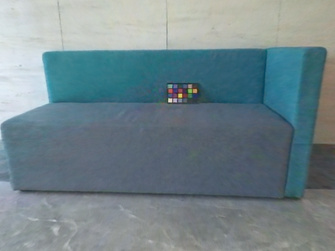} & \includegraphics[width=0.162\textwidth]{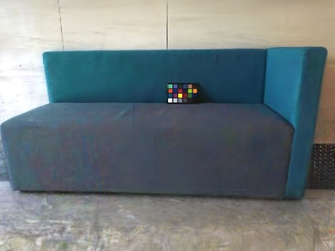} \\
\small Low / Normal light & \small Luminance-GS~\cite{Cui2025LuminanceGS} & \small Aleth-NeRF~\cite{Cui2024AlethNeRF} &
\small LITA-GS~\cite{Zhou2025LITAGS} & \small I\textsuperscript{2}-NeRF~\cite{Liu2025I2NeRF} & \small VeilGS (ours) \\
\end{tabular}
\caption{\textbf{Qualitative comparison on LOM}: \textsc{bike} and \textsc{sofa}; first
column: low light (lower left) and normal light (upper right).}
\label{fig:qual_lom}
\end{figure*}

\begin{table*}[t]
\centering\scriptsize
\setlength{\tabcolsep}{2.2pt}
\caption{\textbf{Restoration on LOM}, against the normal-light frames. ``A + B'': B
enhances the renders of A; ``B + A'': B enhances the training images of A. MOS: user
study. Red: best; yellow: second best.}
\label{tab:lom}
\resizebox{\textwidth}{!}{\begin{tabular}{l|ccc|ccc|ccc|ccc|ccc|cccc}
\hline
Method & \multicolumn{3}{c|}{\textsc{bike}} & \multicolumn{3}{c|}{\textsc{buu}} & \multicolumn{3}{c|}{\textsc{chair}} & \multicolumn{3}{c|}{\textsc{shrub}} & \multicolumn{3}{c|}{\textsc{sofa}} & \multicolumn{4}{c}{Mean} \\
 & PSNR$\uparrow$ & SSIM$\uparrow$ & LPIPS$\downarrow$ & PSNR$\uparrow$ & SSIM$\uparrow$ & LPIPS$\downarrow$ & PSNR$\uparrow$ & SSIM$\uparrow$ & LPIPS$\downarrow$ & PSNR$\uparrow$ & SSIM$\uparrow$ & LPIPS$\downarrow$ & PSNR$\uparrow$ & SSIM$\uparrow$ & LPIPS$\downarrow$ & PSNR$\uparrow$ & SSIM$\uparrow$ & LPIPS$\downarrow$ & MOS$\uparrow$ \\
\hline
NeRF~\cite{Mildenhall2021NeRF} & 6.35 & 0.072 & 0.623 & 7.51 & 0.291 & 0.448 & 6.04 & 0.147 & 0.594 & 8.03 & 0.031 & 0.680 & 6.28 & 0.210 & 0.568 & 6.84 & 0.150 & 0.583 & -- \\
3DGS~\cite{Kerbl2023} & 6.38 & 0.071 & 0.822 & 7.74 & 0.292 & 0.459 & 6.26 & 0.146 & 0.761 & 8.74 & 0.039 & 0.604 & 6.21 & 0.201 & 0.918 & 7.07 & 0.150 & 0.713 & -- \\
\hline
NeRF + SCI~\cite{Ma2022SCI} & 18.84 & 0.637 & 0.565 & 7.84 & 0.660 & 0.562 & 12.07 & 0.699 & 0.584 & 12.38 & 0.358 & 0.587 & 10.25 & 0.737 & 0.626 & 12.28 & 0.618 & 0.585 & -- \\
3DGS + SCI & 8.93 & 0.364 & 0.554 & 18.33 & 0.869 & 0.184 & 11.51 & 0.631 & 0.406 & 12.63 & 0.382 & 0.277 & 12.98 & 0.709 & 0.603 & 12.88 & 0.591 & 0.405 & -- \\
NeRF + Zero-DCE~\cite{Guo2020ZeroDCE} & 10.16 & 0.468 & 0.557 & 17.81 & 0.833 & 0.357 & 12.44 & 0.684 & 0.547 & 12.58 & 0.282 & 0.540 & 14.43 & 0.787 & 0.539 & 13.48 & 0.611 & 0.508 & -- \\
Zero-DCE + NeRF & 10.39 & 0.518 & 0.464 & 17.90 & 0.858 & 0.376 & 12.58 & 0.721 & 0.460 & 12.32 & 0.308 & 0.481 & 14.45 & 0.831 & 0.419 & 13.53 & 0.647 & 0.440 & -- \\
Zero-DCE + 3DGS & 10.54 & 0.539 & 0.401 & 17.92 & \ranksecond{0.896} & \ranksecond{0.179} & 12.94 & 0.756 & 0.303 & 13.10 & 0.467 & 0.229 & 14.42 & 0.831 & 0.356 & 13.78 & 0.698 & 0.294 & -- \\
SCI + NeRF & 13.44 & 0.658 & 0.435 & 7.76 & 0.692 & 0.525 & 19.77 & 0.802 & 0.674 & 18.16 & 0.503 & 0.475 & 10.08 & 0.772 & 0.520 & 13.84 & 0.685 & 0.526 & -- \\
3DGS + Zero-DCE & 10.56 & 0.498 & 0.500 & 18.86 & 0.890 & 0.191 & 13.24 & 0.731 & 0.349 & 13.26 & 0.430 & 0.272 & 14.23 & 0.767 & 0.586 & 14.03 & 0.663 & 0.380 & -- \\
SCI + 3DGS & 13.67 & 0.677 & 0.324 & 7.95 & 0.695 & 0.501 & 21.77 & 0.866 & 0.350 & 18.67 & 0.657 & \rankfirst{0.153} & 9.99 & 0.750 & 0.452 & 14.41 & 0.729 & 0.356 & -- \\
\hline
Luminance-GS~\cite{Cui2025LuminanceGS} & 18.27 & 0.749 & 0.411 & 18.09 & 0.877 & 0.193 & 19.82 & 0.835 & 0.367 & 15.40 & \ranksecond{0.666} & 0.241 & 20.12 & 0.871 & 0.259 & 18.34 & 0.800 & 0.294 & 2.23 \\
Aleth-NeRF~\cite{Cui2024AlethNeRF} & 20.46 & 0.727 & 0.499 & 20.22 & 0.859 & 0.315 & 20.93 & 0.818 & 0.468 & 18.24 & 0.511 & 0.448 & 19.52 & 0.857 & 0.354 & 19.87 & 0.754 & 0.417 & 3.27 \\
LITA-GS~\cite{Zhou2025LITAGS} & 22.75 & \rankfirst{0.819} & \ranksecond{0.282} & 20.59 & \rankfirst{0.897} & \rankfirst{0.175} & 22.60 & \ranksecond{0.873} & \ranksecond{0.223} & \ranksecond{19.35} & 0.659 & 0.217 & 20.43 & \rankfirst{0.895} & 0.268 & 21.14 & \rankfirst{0.829} & \ranksecond{0.233} & 3.11 \\
I\textsuperscript{2}-NeRF~\cite{Liu2025I2NeRF} & \ranksecond{22.87} & 0.803 & \rankfirst{0.278} & \ranksecond{22.35} & 0.845 & 0.203 & \rankfirst{23.82} & 0.847 & 0.280 & \rankfirst{19.44} & \rankfirst{0.674} & 0.249 & \ranksecond{25.89} & 0.859 & \ranksecond{0.234} & \ranksecond{22.87} & 0.806 & 0.249 & \ranksecond{3.38} \\
VeilGS (ours) & \rankfirst{23.20} & \ranksecond{0.815} & 0.287 & \rankfirst{22.52} & 0.893 & 0.181 & \ranksecond{23.65} & \rankfirst{0.884} & \rankfirst{0.216} & 19.21 & 0.662 & \ranksecond{0.205} & \rankfirst{26.21} & \ranksecond{0.887} & \rankfirst{0.212} & \rankfirst{22.96} & \ranksecond{0.828} & \rankfirst{0.220} & \rankfirst{3.57} \\
\hline
\end{tabular}
}
\end{table*}

\subsection{Training Objectives}
\label{sec:losses}

We train the Gaussians, the shading MLP, the basis network and the physical biases
end to end. Only the reconstruction loss compares $\hat{I}$ with $I$; the other three losses
are priors, and none of them compares $J$ with a reference image.

\noindent\textbf{Reconstruction.}
Following 3DGS~\cite{Kerbl2023},
\begin{equation}
  \mathcal{L}_{\mathrm{rec}}=(1-\lambda_{\mathrm{s}})\,\|\hat{I}-I\|_1
  +\lambda_{\mathrm{s}}\big(1-\mathrm{SSIM}(\hat{I},I)\big),
  \label{eq:rec}
\end{equation}
with $\lambda_{\mathrm{s}}=0.2$. This loss sees $J$ only through the veil, so it cannot
fix the brightness or the albedo--shading split of $J$, and fixes its colour only up
to $\gamma$.

\noindent\textbf{Colour correction.}
Two priors set the brightness that $\mathcal{L}_{\mathrm{rec}}$ leaves open and
balance the colour,
\begin{equation}
  \mathcal{L}_{\mathrm{cc}}=\lambda_{\mathrm{exp}}\,\mathcal{L}_{\mathrm{exp}}
  +\lambda_{\mathrm{gw}}\,\mathcal{L}_{\mathrm{gw}},
  \label{eq:colourcorr}
\end{equation}
with $\lambda_{\mathrm{exp}}=0.3$ and $\lambda_{\mathrm{gw}}=0.05$. Inspired by
Zero-DCE~\cite{Guo2020ZeroDCE}, the exposure term pulls the mean of $J$ toward a target
$E=0.4$, as in RestorGS~\cite{Qiao2025RestorGS},
\begin{equation}
  \mathcal{L}_{\mathrm{exp}}=\Big|\frac{1}{3|\Omega|}\sum_{\p\in\Omega}
  \sum_{\ell\in\{R,G,B\}}J(\p,\ell)\;-\;E\Big|,
  \label{eq:exposure}
\end{equation}
where $\Omega$ is the set of pixels.
Since a dark capture gives no reliable colour reference, the gray-world term follows
the colour constancy loss of Zero-DCE and asks the channel means of $J$ to be equal,
\begin{equation}
  \mathcal{L}_{\mathrm{gw}}=\!\!\sum_{(\ell,\ell')\in\mathcal{P}}\!\!
  \big(\bar{J}_{\ell}-\bar{J}_{\ell'}\big)^2,\quad
  \bar{J}_{\ell}=\frac{1}{|\Omega|}\sum_{\p\in\Omega}J(\p,\ell),
  \label{eq:grayworld}
\end{equation}
where $\mathcal{P}=\{(R,G),(G,B),(R,B)\}$.

\noindent\textbf{Retinex.}
Following Retinex theory~\cite{Land1977Retinex}, we constrain each factor of
Eq.~\eqref{eq:decomp},
\begin{equation}
  \mathcal{L}_{\mathrm{retinex}}=\lambda_{\mathrm{alb}}\,\mathcal{L}_{\mathrm{alb}}
  +\lambda_{\mathrm{sh}}\,\mathcal{L}_{\mathrm{sh}},
  \label{eq:retinex}
\end{equation}
with $\lambda_{\mathrm{alb}}=0.5$ and $\lambda_{\mathrm{sh}}=0.02$. The albedo should lie
in $[0,1]$ and be similar across neighbouring Gaussians,
\begin{equation}
  \mathcal{L}_{\mathrm{alb}}=\mathbb{E}_i\Big[\rho_{[0,1]}(a_i)
  +\mu_{a}\!\!\sum_{i'\in\mathcal{N}(i)}\!\!w_{ii'}\,\|a_i-a_{i'}\|^2\Big],
  \label{eq:albedo}
\end{equation}
where $\mathcal{N}(i)$ holds the 8 nearest Gaussians in 3D, $w_{ii'}$ decays with
distance, $\rho_{[l,h]}$ is zero inside $[l,h]$ and linear outside, and $\mu_a=0.01$.
The shading should be grey, so that colour lives in the albedo, and lie between
$s_{\min}=0.1$ and $s_{\max}=1.5$:
\begin{equation}
  \mathcal{L}_{\mathrm{sh}}=\mathbb{E}_i\Big[\|s_i-\bar{s}_i\mathbf{1}\|^2
  +\mu_{s}\,\rho_{[s_{\min},s_{\max}]}(s_i)^2\Big],
  \label{eq:shading}
\end{equation}
where $\bar{s}_i$ is the mean of the three channels of $s_i$ and $\mathbf{1}$ is the
all-ones vector, so that $\bar{s}_i\mathbf{1}$ is a grey shading with the same mean,
and $\mu_s=2.5$.

\noindent\textbf{Depth.}
The inverse depth $D$, blended from per-Gaussian depths as in Eq.~\eqref{eq:blend},
follows a monocular estimate $\tilde{D}$ from Depth
Anything~3~\cite{Lin2025DepthAnything3} on the dark images, both normalized to
$[0,1]$~\cite{Kerbl2024Hierarchical},
\begin{equation}
  \mathcal{L}_{\mathrm{depth}}=\|D-\tilde{D}\|_1 .
  \label{eq:depth}
\end{equation}

\noindent\textbf{Total loss.}
The total loss is
\begin{equation}
  \mathcal{L}=\mathcal{L}_{\mathrm{rec}}+\mathcal{L}_{\mathrm{cc}}
  +\mathcal{L}_{\mathrm{retinex}}+\lambda_{\mathrm{d}}\,\mathcal{L}_{\mathrm{depth}},
  \label{eq:total}
\end{equation}
with $\lambda_{\mathrm{d}}$ decaying from 0.2 to 0.01 during training. All weights,
including $E$, are the same for every scene and both benchmarks, and no
hyperparameter was tuned on normal-light frames.

\section{Experiments}
\label{sec:experiments}

\subsection{Experimental Setup}
\label{sec:ExpSetup}

\noindent\textbf{Evaluation Datasets and Metrics.}
We evaluate on two real low-light benchmarks. The Gaussian-DK night benchmark~\cite{Ye2024GaussianDK} has twelve scenes
of \mbox{75--131} dark views with varying exposure and no normal-light reference. We
train at quarter resolution and hold out \mbox{9--15} views per scene. We score the
restored view, which for VeilGS is $J$ rendered from the camera pose alone, with the
no-reference metrics NIQE~\cite{Mittal2013NIQE}, MUSIQ~\cite{Ke2021MUSIQ} and
CLIP-IQA~\cite{Wang2023CLIPIQA}. As there is no normal-light reference, Gaussian-DK and
RestorGS~\cite{Qiao2025RestorGS} evaluate reconstruction with PSNR,
SSIM~\cite{Wang2004SSIM} and LPIPS~\cite{Zhang2018LPIPS} between each held-out dark view
and its dark rendering. For VeilGS this reconstruction is target-conditioned: it
computes the basis and the coefficients of the veil from the held-out view, with all
trained parameters frozen. We therefore use it mainly for the ablations and give the
comparison with other methods in the appendix (Table~\ref{tab:sup_dk_recon}).
LOM~\cite{Cui2024AlethNeRF} has five
scenes (\textsc{bike}, \textsc{buu}, \textsc{chair}, \textsc{shrub}, \textsc{sofa}),
each captured with paired low-light and normal-light frames at $500\times375$. We
train on the low-light images and, on the official test split of \mbox{3--5} views per
scene, compare $J$ with the normal-light frames by PSNR, SSIM and LPIPS, without using
the low-light test views. To get human feedback, we also ran a user study in which 55 participants rated
the restored renders of VeilGS and four other methods on both benchmarks, blind and on
a 1--5 scale, and we report the mean opinion score (details, consent and ethics in
Sec.~\ref{sup:study}).

\noindent\textbf{Implementation Details.}
We build on the official 3DGS implementation~\cite{Kerbl2023} and train each scene for
35{,}000 iterations on a single NVIDIA H100 GPU, with densification until iteration
15{,}000 and a densification gradient threshold of $10^{-4}$. Training, including
periodic evaluation, takes 20--22 minutes per scene on LOM and 36--68 minutes on
Gaussian-DK. On LOM, VeilGS averages 20.7 minutes and renders $J$ at 407 FPS, against
5.1 minutes and 397 FPS for Luminance-GS on the same GPU and resolution
(Fig.~\ref{fig:teaser} and Table~\ref{tab:sup_efficiency}). The shading MLP $f_{\theta}$ has two hidden layers of width 128. Camera
poses and SfM points come from the COLMAP reconstruction released with each
benchmark; albedo is initialized to the colours of the SfM points and the shading
latents to zero.
Sec.~\ref{sup:impl} gives the training schedule.

\subsection{Quantitative and Qualitative Comparison}
\label{sec:Comp}

\noindent\textbf{Baselines.}
On Gaussian-DK we compare with 3DGS~\cite{Kerbl2023}, Gaussian-DK~\cite{Ye2024GaussianDK},
LITA-GS~\cite{Zhou2025LITAGS} and Luminance-GS~\cite{Cui2025LuminanceGS}, each trained
with its official code on the same training split and evaluated on the same held-out
views; the appendix (Table~\ref{tab:sup_dk_recon}) adds the reported results of
NeRF-W~\cite{MartinBrualla2021NeRFW} and RestorGS~\cite{Qiao2025RestorGS}. On LOM we compare with NeRF~\cite{Mildenhall2021NeRF}, 3DGS,
their combinations with Zero-DCE~\cite{Guo2020ZeroDCE} and SCI~\cite{Ma2022SCI},
Aleth-NeRF~\cite{Cui2024AlethNeRF}, Luminance-GS~\cite{Cui2025LuminanceGS}, LITA-GS~\cite{Zhou2025LITAGS} and
I\textsuperscript{2}-NeRF~\cite{Liu2025I2NeRF}, using the results reported
in~\cite{Zhou2025LITAGS,Cui2025LuminanceGS,Liu2025I2NeRF}.

\noindent\textbf{Gaussian-DK.}
This benchmark has no clean ground truth, so we assess restoration with no-reference
metrics and the user study. Table~\ref{tab:dk_noref} scores the restored renders.
VeilGS has the best NIQE on all twelve scenes and the best CLIP-IQA on ten, and is second in MUSIQ,
where Gaussian-DK~\cite{Ye2024GaussianDK} and VeilGS are each best on six scenes.
Plain 3DGS scores above LITA-GS and Luminance-GS on all three metrics and in MOS, likely
because their over-exposure and haze (Fig.~\ref{fig:qual_dk}) cost more than its
darkness. In Fig.~\ref{fig:qual_dk}, Gaussian-DK and
Luminance-GS leave a grey haze over the scene, most visible on the walls and floor of
\textsc{dormitory} and \textsc{kitchen}, and LITA-GS~\cite{Zhou2025LITAGS} over-exposes the surfaces near light
sources, such as the lantern in \textsc{tree} and the wall in \textsc{street}. VeilGS
lifts the dark regions without saturating the lit ones, so the scene keeps its contrast
between lit and unlit areas.

\noindent\textbf{LOM.}
Table~\ref{tab:lom} compares the restored renders with the normal-light frames. VeilGS
is competitive, with modest leads in mean PSNR and LPIPS: 22.96\,dB against 22.87\,dB
for I\textsuperscript{2}-NeRF~\cite{Liu2025I2NeRF}, and 0.220 against 0.233 for
LITA-GS~\cite{Zhou2025LITAGS}. Its SSIM is second, 0.001 below LITA-GS, and it has the
highest PSNR on three of the five scenes. In Fig.~\ref{fig:qual_lom}, Luminance-GS~\cite{Cui2025LuminanceGS} washes
out the colours and Aleth-NeRF~\cite{Cui2024AlethNeRF} blurs fine structures such as the wall tiles in \textsc{bike}, while
VeilGS, LITA-GS and I\textsuperscript{2}-NeRF recover sharp details close to the
normal-light frames.

\noindent\textbf{User study.}
According to the blind user study, VeilGS has the highest mean opinion score (MOS):
4.15 against 3.44 for Gaussian-DK~\cite{Ye2024GaussianDK} (Table~\ref{tab:dk_noref}),
and 3.57 against 3.38 for I\textsuperscript{2}-NeRF~\cite{Liu2025I2NeRF} on LOM
(Table~\ref{tab:lom}). Both differences are significant in a two-sided Wilcoxon
signed-rank test over participants ($p<10^{-9}$ and $p=0.001$; Sec.~\ref{sup:study}).
On \textsc{piano} and \textsc{snowman}, however, participants preferred Gaussian-DK,
although VeilGS has better NIQE, MUSIQ and CLIP-IQA there; these metrics appear to
miss local colour artefacts in its renders.

\subsection{Ablation Results}
\label{sec:Ablation}

We ablate VeilGS on Gaussian-DK in two parts: how the veil coefficients are obtained
(Table~\ref{tab:abl_mech}), and what each component of the model contributes
(Table~\ref{tab:abl_comp}). Both tables report the target-conditioned reconstruction of
Sec.~\ref{sec:ExpSetup} and, besides NIQE and MUSIQ, the perceptual index
(PI)~\cite{Blau2018PIRM}.

\begin{table}[t]
\centering\scriptsize
\setlength{\tabcolsep}{2pt}
\caption{\textbf{How the veil coefficients are obtained}, over the twelve
Gaussian-DK scenes. Red: best; yellow: second best.}
\label{tab:abl_mech}
\resizebox{\columnwidth}{!}{\begin{tabular}{l|ccc|ccc}
\hline
 & \multicolumn{3}{c|}{Reconstruction} & \multicolumn{3}{c}{No reference} \\
Variant & PSNR$\uparrow$ & SSIM$\uparrow$ & LPIPS$\downarrow$ & NIQE$\downarrow$ & MUSIQ$\uparrow$ & PI$\downarrow$ \\
\hline
Learned per-view table & 19.62 & 0.631 & 0.170 & 3.06 & 62.94 & 3.11 \\
Additive basis, detached & 24.74 & 0.798 & 0.122 & 2.90 & 65.37 & 2.98 \\
Rank-one basis, detached & 25.04 & 0.816 & 0.119 & 2.89 & 65.83 & 3.04 \\
Additive basis, through & \ranksecond{25.96} & \ranksecond{0.832} & \rankfirst{0.103} & \rankfirst{2.62} & \ranksecond{67.92} & \rankfirst{2.80} \\
\hline
Rank-one basis, through (ours) & \rankfirst{26.39} & \rankfirst{0.841} & \ranksecond{0.104} & \ranksecond{2.63} & \rankfirst{68.37} & \rankfirst{2.80} \\
\hline
\end{tabular}
}
\end{table}

\begin{table}[t]
\centering\scriptsize
\setlength{\tabcolsep}{2pt}
\caption{\textbf{Components of VeilGS}, over the four scenes of
Fig.~\ref{fig:qual_dk}, two indoor and two outdoor, sampled to limit training cost.
Each row removes one part of the full model. Red: best; yellow: second best.}
\label{tab:abl_comp}
\resizebox{\columnwidth}{!}{\begin{tabular}{l|ccc|ccc}
\hline
 & \multicolumn{3}{c|}{Reconstruction} & \multicolumn{3}{c}{No reference} \\
Variant & PSNR$\uparrow$ & SSIM$\uparrow$ & LPIPS$\downarrow$ & NIQE$\downarrow$ & MUSIQ$\uparrow$ & PI$\downarrow$ \\
\hline
w/o spatially varying veil & 21.79 & 0.688 & 0.166 & 2.80 & 65.47 & 2.96 \\
w/o damping & 21.91 & 0.654 & 0.336 & 3.71 & 52.53 & 3.79 \\
spherical harmonics colour & 24.23 & 0.802 & 0.137 & 3.94 & 66.96 & 2.92 \\
w/o Retinex terms & 25.69 & 0.813 & 0.114 & 2.67 & 68.60 & 2.92 \\
w/o depth prior & \ranksecond{26.35} & \ranksecond{0.829} & \ranksecond{0.109} & \ranksecond{2.66} & \ranksecond{69.63} & \ranksecond{2.84} \\
\hline
Full model & \rankfirst{26.73} & \rankfirst{0.836} & \rankfirst{0.107} & \rankfirst{2.56} & \rankfirst{70.25} & \rankfirst{2.76} \\
\hline
\end{tabular}
}
\end{table}

\begin{figure}[t]
\centering
\setlength{\tabcolsep}{0.5pt}
\renewcommand{\arraystretch}{0.4}
\begin{tabular}{ccc}
\includegraphics[width=0.325\columnwidth]{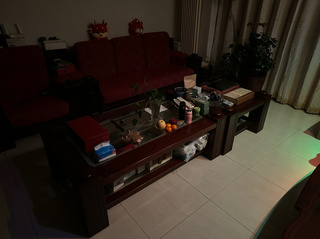} & \includegraphics[width=0.325\columnwidth]{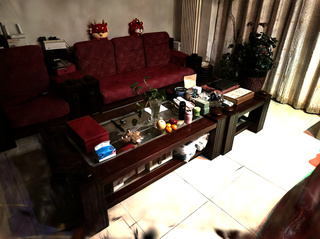} & \includegraphics[width=0.325\columnwidth]{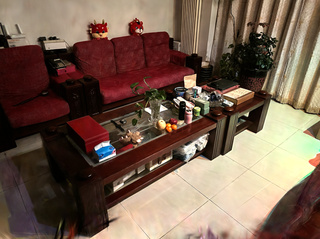} \\
\small (a) & \small (b) & \small (c) \\
\includegraphics[width=0.325\columnwidth]{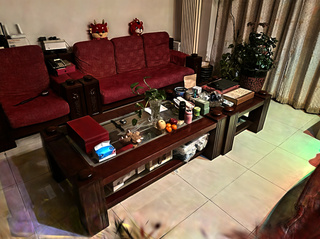} & \includegraphics[width=0.325\columnwidth]{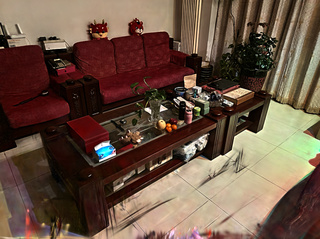} & \includegraphics[width=0.325\columnwidth]{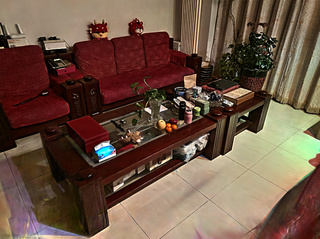} \\
\small (d) & \small (e) & \small (f) \\
\end{tabular}
\caption{\textbf{How the veil coefficients are obtained}, on a held-out view of
\textsc{livingroom}. (a) the dark observation; the restored field with (b) a learned
per-view coefficient table, (c) an additive basis detached, (d) a rank-one basis detached,
(e) an additive basis through the solve and (f) our rank-one basis through the solve.}
\label{fig:abl_mech_qual}
\end{figure}

\noindent\textbf{How the coefficients are obtained.}
Table~\ref{tab:abl_mech} compares ways of obtaining the veil coefficients, and
Fig.~\ref{fig:abl_mech_qual} shows them on a held-out view. Learning a
coefficient table per training view is the weakest on every metric. A held-out view has
no entry in the table, so it uses the average of the learned coefficients, and its PSNR,
6.77\,dB below our model, partly reflects this; its no-reference scores, which depend
only on $J$, are also the worst. Training through the solve improves on detaching its gradient for both bases
and on all six metrics: by 1.35\,dB PSNR and 2.5 MUSIQ with the rank-one basis and by
1.22\,dB and 2.6 MUSIQ with the additive one, which is consistent with the
argument of Sec.~\ref{sec:varpro}. Rank-one is ahead of additive
by 0.43\,dB PSNR through the solve and by 0.30\,dB when detaching it. Our configuration,
rank-one trained through the solve, is best in PSNR, SSIM and MUSIQ,
ties on PI, and is within 0.01 of the additive basis in NIQE and LPIPS.

\begin{figure}[t]
\centering
\setlength{\tabcolsep}{0.5pt}
\renewcommand{\arraystretch}{0.4}
\begin{tabular}{cccc}
\includegraphics[width=0.2435\columnwidth]{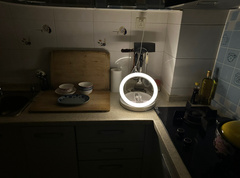} & \includegraphics[width=0.2435\columnwidth]{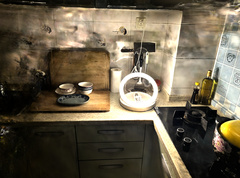} & \includegraphics[width=0.2435\columnwidth]{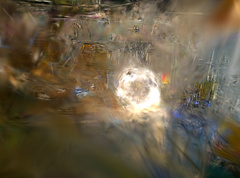} &
\includegraphics[width=0.2435\columnwidth]{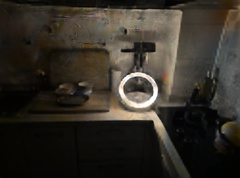} \\
\small (a) & \small (b) & \small (c) & \small (d) \\
\end{tabular}\\[1pt]
\begin{tabular}{cccc}
\includegraphics[width=0.2435\columnwidth]{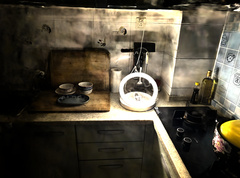} & \includegraphics[width=0.2435\columnwidth]{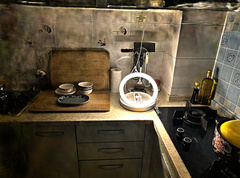} & \includegraphics[width=0.2435\columnwidth]{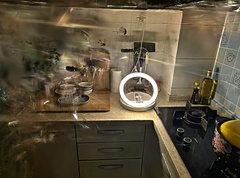} & \includegraphics[width=0.2435\columnwidth]{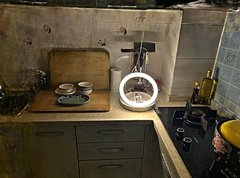} \\
\small (e) & \small (f) & \small (g) & \small (h) \\
\end{tabular}
\caption{\textbf{Components of VeilGS} on a held-out view of \textsc{kitchen}.
(a) the dark observation; the restored field (b) w/o spatially varying veil,
(c) w/o damping, (e) with spherical harmonics colour, (f) w/o Retinex terms, (g) w/o
depth prior and (h) of the full model; (d) the prediction $\hat{I}$ of the undamped
model, \ie (c) under its veil.}
\label{fig:abl_qual}
\end{figure}

\noindent\textbf{Components.}
Table~\ref{tab:abl_comp} removes one part at a time, and every removal worsens all six
metrics; Fig.~\ref{fig:abl_qual} shows the same variants on a held-out view. Solving a
single gain per view costs 4.94\,dB PSNR, and removing the damping of the solve costs
4.82\,dB. Fig.~\ref{fig:abl_qual}(c,\,d) shows that the undamped veil absorbs scene
detail instead of illumination, so the prediction still reproduces the observation
while the restored field has dissolved.
Replacing the albedo and shading by spherical harmonics costs 2.50\,dB PSNR and raises
NIQE by 1.38, the largest rise in NIQE, and leaves the shadows dark while
over-brightening the lit areas (Fig.~\ref{fig:abl_qual}(e)). Removing the Retinex
terms costs 1.04\,dB. Without the depth prior, PSNR drops by 0.38\,dB and MUSIQ by 0.6.

\section{Conclusion}
\label{sec:conclusion}
We presented VeilGS, which treats low-light degradation as a veil drawn over a well-lit
3D Gaussian scene. Each Gaussian carries an albedo and a shading latent instead of
spherical harmonics, and each view is darkened by a spatially varying gain, spanned by a
coarse basis read off the observation, followed by a global tone curve. Rather than
learning parameters for every view, VeilGS solves the veil from each observation and trains the
scene through that solve, which discourages the Gaussians from absorbing what the
veil can explain. On Gaussian-DK, its restored renders give the best NIQE and CLIP-IQA
and are rated highest in our user study; on LOM, it achieves results competitive with
the state of the art. However, the veil
is a single achromatic gain, so it cannot remove colour casts from coloured light
sources. Training through the solve can also become unstable on low-texture scenes.
Finally, the albedo--shading decomposition is imperfect and can leave a green
cast in the albedo, mostly in dark regions.

\section*{LLM Usage Statement}
We used Claude Opus 5.5 (Anthropic) to assist with running the baseline methods, rendering
result videos and editing code, and to polish the writing of the manuscript and its
appendix. We also used it to format tables, compute summary statistics from
our evaluation and user-study results, and write the script that draws
the plot in Fig.~\ref{fig:teaser}. The authors reviewed all outputs and take full responsibility for
the content of this work.

{
    \small
    \bibliographystyle{ieeenat_fullname}
    \bibliography{main}
}

\maketitlesupplementary
\pdfbookmark[1]{Supplementary Material}{supplementary}

\setcounter{section}{0}
\setcounter{figure}{0}
\setcounter{table}{0}
\setcounter{equation}{0}
\renewcommand{\thesection}{S\arabic{section}}
\renewcommand{\theequation}{S\arabic{equation}}
\renewcommand{\thefigure}{S\arabic{figure}}
\renewcommand{\thetable}{S\arabic{table}}
\renewcommand{\theHsection}{S\arabic{section}}
\renewcommand{\theHequation}{S\arabic{equation}}
\renewcommand{\theHfigure}{S\arabic{figure}}
\renewcommand{\theHtable}{S\arabic{table}}

This supplementary material gives details and results that complement the main paper.
Secs.~\ref{sup:formation}--\ref{sup:varpro} describe the image formation model, the
veil, the solver and the effect of training through the solve. Sec.~\ref{sup:impl}
gives further implementation details, Sec.~\ref{sup:results} the evaluation protocol
and additional results, and Sec.~\ref{sup:study} the user study. Numbers without an
``S'' refer to the main paper.

\section{Low-Light Image Formation}
\label{sup:formation}

\noindent\textbf{Tone curve.}
Eq.~\eqref{eq:lowlight} applies the camera's tone curve to each colour channel of
$x=\alpha J$. In practice, we average this per-channel curve with a curve on the
luminance,
\begin{equation}
  \hat{I}(\p)=\eta\,x(\p)\,Y\big(x(\p)\big)^{\frac{1}{\gamma}-1}
               +(1-\eta)\,x(\p)^{\frac{1}{\gamma}}.
  \label{eq:sup_blend}
\end{equation}
The second term is the per-channel curve of Eq.~\eqref{eq:lowlight}. The first term
applies the curve to the luminance $Y(x)=0.299\,x_R+0.587\,x_G+0.114\,x_B$ and keeps
the ratios between the channels, following Schlick~\cite{Schlick1994}. A per-channel
curve fades colours~\cite{Mantiuk2009Color}, while a ratio-preserving curve can shift
hue. We therefore average the two, with $\eta=0.5$ in all experiments. The forward
model clips $\alpha$ to $[10^{-3},1]$, $\gamma$ to $[1,4]$ and $\hat{I}$ to $[0,1]$.

\noindent\textbf{Linearity in the veil.}
The veil $\alpha$ is one number per pixel, and $Y$ is linear, so
$Y(\alpha J)=\alpha\,Y(J)$. Hence $\alpha^{1/\gamma}$ factors out of both terms of
Eq.~\eqref{eq:sup_blend}:
\begin{equation}
  \hat{I}(\p)=\alpha(\p)^{1/\gamma}\Big[\eta\,J\,Y(J)^{\frac{1}{\gamma}-1}
  +(1-\eta)\,J^{\frac{1}{\gamma}}\Big](\p).
  \label{eq:sup_factor}
\end{equation}
With $\alpha=b_{\alpha}e^{m}$, the log of Eq.~\eqref{eq:sup_factor} at pixel $\p$ and
channel $\ell$ is exactly
\begin{equation}
  \log\hat{I}=\tfrac{1}{\gamma}\log b_{\alpha}
  +\log\Big[\eta\,J\,Y(J)^{\frac{1}{\gamma}-1}+(1-\eta)J^{\frac{1}{\gamma}}\Big]
  +\tfrac{1}{\gamma}m.
  \label{eq:sup_logblend}
\end{equation}
The veil enters only through $\frac{1}{\gamma}m$. For $\eta=0$,
Eq.~\eqref{eq:sup_logblend} becomes Eq.~\eqref{eq:loglin} of the main paper,
\begin{equation}
  \log\hat{I}(\p,\ell)=\tfrac{1}{\gamma}\big(\log b_{\alpha}+\log J(\p,\ell)+m(\p)\big),
  \label{eq:sup_loglin}
\end{equation}
which is the relation the solver uses. The blend changes only the part that does not
depend on the veil. The slope $\frac{1}{\gamma}$ of $m$, and therefore the form of the
fit, stay the same.

\section{The Veil}
\label{sup:veil}

\noindent\textbf{Basis.}
The basis network $h_{\phi}$ has a stem convolution, four residual blocks of width 32
and an output head. It runs at one-eighth resolution, and its output is enlarged
bilinearly, so each map has 64 times fewer degrees of freedom than the image. A $\tanh$
keeps every map in $[-1,1]$, so the scale clip of Sec.~\ref{sup:solver} bounds the
modulation $m$. Tang~\etal~\cite{Tang2022ZIC} combine their two bases by convolution.
We multiply them pixel by pixel in Eq.~\eqref{eq:bilinear}, since the veil is a
smooth brightness field rather than a blur. The factors $(\uu,\vv,s)$ are nine numbers,
of which seven are free, because $\uu$ and $\vv$ have unit length.

\begin{figure*}[t]
\centering
\setlength{\tabcolsep}{1pt}
\renewcommand{\arraystretch}{0.5}
\begin{tabular}{ccccccc}
\includegraphics[width=0.138\textwidth]{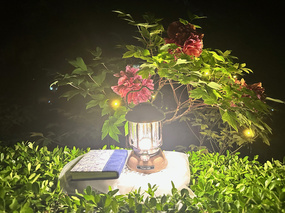} & \includegraphics[width=0.138\textwidth]{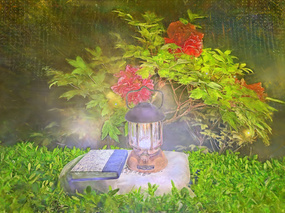} & \includegraphics[width=0.138\textwidth]{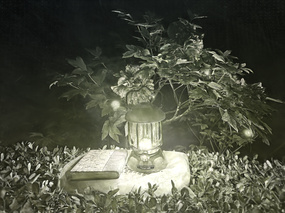} &
\includegraphics[width=0.138\textwidth]{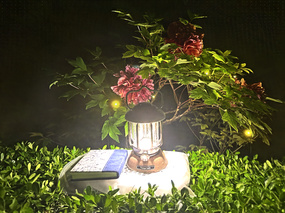} & \includegraphics[width=0.138\textwidth]{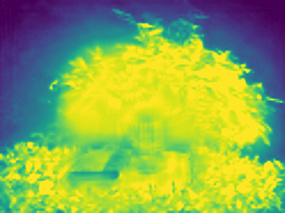} & \includegraphics[width=0.138\textwidth]{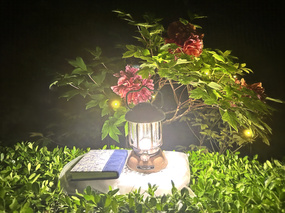} &
\includegraphics[width=0.138\textwidth]{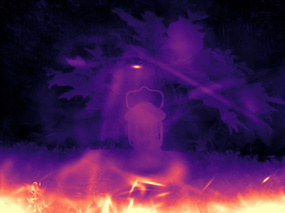} \\
\includegraphics[width=0.138\textwidth]{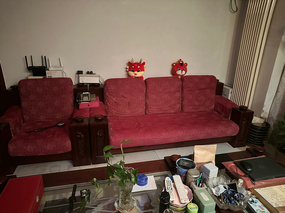} & \includegraphics[width=0.138\textwidth]{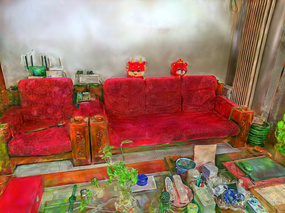} & \includegraphics[width=0.138\textwidth]{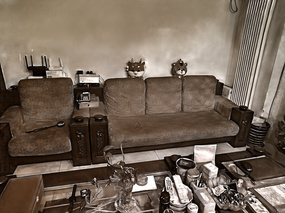} &
\includegraphics[width=0.138\textwidth]{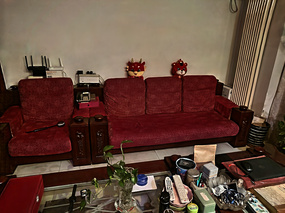} & \includegraphics[width=0.138\textwidth]{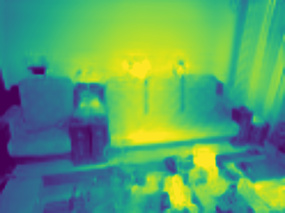} & \includegraphics[width=0.138\textwidth]{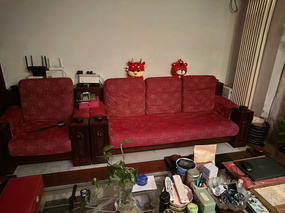} &
\includegraphics[width=0.138\textwidth]{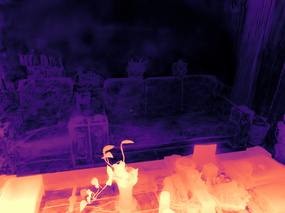} \\
\small Observation $I$ & \small Albedo & \small Shading &
\small Restored $J$ & \small Veil $\alpha$ & \small Prediction $\hat{I}$ &
\small Inverse depth \\
\end{tabular}
\caption{\textbf{What VeilGS stores and renders}, on training views of \textsc{flower}
(top) and \textsc{livingroom} (bottom). The albedo $a_i$ and the shading $s_i$ are each
rasterized like a colour, with the shading divided by its 99th percentile, so their
product only approximates $J$. The veil is solved from the observation and shown from
its 1st to its 99th percentile, as in Fig.~\ref{fig:teaser}. The prediction applies
the tone curve of Eq.~\eqref{eq:sup_blend} to $\alpha J$.}
\label{fig:sup_decomp}
\end{figure*}

\noindent\textbf{What each part holds.}
Fig.~\ref{fig:sup_decomp} shows the intermediate quantities on two training views. The
shading is close to grey, as Eq.~\eqref{eq:shading} asks, so the colour sits in the
albedo. The shading is below one over most of the image (median 0.29 on
\textsc{flower} and 0.45 on \textsc{livingroom}), so the albedo is brighter and more
saturated than $J$. The veil, solved from the observation as in training, is highest
where light falls and lowest in unlit regions. Under this veil, the prediction
$\hat{I}$ matches the observation at 26.2\,dB (\textsc{flower}) and 29.3\,dB
(\textsc{livingroom}) PSNR.

\section{Solver Details}
\label{sup:solver}

\noindent\textbf{Problem.}
Substituting Eq.~\eqref{eq:bilinear} into Eq.~\eqref{eq:sup_loglin} and stacking all
pixels and channels gives the fit of Eq.~\eqref{eq:ls} in matrix form,
\begin{equation}
  \min_{s,\uu,\vv}\;\big\|A\,\vecop(M)-r\big\|^2+\vecop(M)^{\top}\Lambda\,\vecop(M),
  \quad M=s\,\uu\vv^{\top}.
  \label{eq:sup_ls}
\end{equation}
Here $r$ stacks the residuals of Eq.~\eqref{eq:residual}, and $\vecop$ stacks the
rows of $M$, so that $\vecop(\uu\vv^{\top})=\uu\otimes\vv$. The design matrix $A$ has
the entries
\begin{equation}
  A_{(\p,\ell),(j,k)}=\tfrac{1}{\gamma}\,\Phi_{jk}(\p)
  =\tfrac{1}{\gamma}\,K_j(\p)\,X_k(\p).
  \label{eq:sup_design}
\end{equation}
Each column is one product map $K_j\odot X_k$, scaled by $\frac{1}{\gamma}$ and repeated
for the three channels. At the $64\times64$ solve resolution, $A$ has
$3\cdot64^2=12{,}288$ rows and 16 columns. The stages below use only $A^{\top}A$
($16\times16$) and $A^{\top}r$.

The three rows of a pixel are equal, so the fit sees $r$ only through its channel mean
$\bar{r}(\p)=\frac{1}{3}\sum_{\ell}r(\p,\ell)$. For any value $q$ of the veil term,
\begin{equation}
  \sum_{\ell}\big(q-r(\p,\ell)\big)^2
  =3\big(q-\bar{r}(\p)\big)^2+\sum_{\ell}\big(r(\p,\ell)-\bar{r}(\p)\big)^2,
  \label{eq:sup_chmean}
\end{equation}
and the last term does not depend on the coefficients. The fit therefore has
$64^2=4{,}096$ distinct equations.

\noindent\textbf{Damping.}
We use $\Lambda=\lambda\operatorname{diag}(A^{\top}A)+\epsilon\mathbf{I}$ with
$\lambda=4$ and $\epsilon=10^{-4}$~\cite{Marquardt1963}. The first term damps each
coefficient in proportion to its own curvature. This keeps the conditioning
independent of the scale of $A$, which changes with $\gamma$ during training. The floor
$\epsilon$ keeps the system invertible if a column of $A$ vanishes. We form
$H=A^{\top}A+\Lambda$ once and use it in all three stages, so they all minimize the same
damped objective.

\noindent\textbf{Stage 1: joint solve.}
Without the rank constraint, the solution is $\vecop(\tilde{M})=H^{-1}A^{\top}r$. We
compute it with a QR factorization of $H$.

\noindent\textbf{Stage 2: rank-one initialization.}
Eight power iterations on $\tilde{M}^{\top}\tilde{M}$ give $\vv$. Then
$\uu=\tilde{M}\vv/\|\tilde{M}\vv\|$ and $s=\uu^{\top}\tilde{M}\vv$, with the sign absorbed
into $\uu$ so that $s\ge0$. This approximates the rank-one matrix closest to $\tilde{M}$
in the Frobenius norm. It is in general not the minimizer of Eq.~\eqref{eq:sup_ls}, so
it only initializes the next stage. Tang~\etal take a single iteration from a random
vector. We start from the all-ones vector instead, which removes a source of randomness
from the training graph.

\noindent\textbf{Stage 3: Gauss--Newton refinement.}
Let $T=[\,\uu\otimes\mathbf{I}_d\;|\;\mathbf{I}_d\otimes\vv\,]$ be the Jacobian of
$\uu\otimes\vv$ with respect to $(\vv,\uu)$. Each of four steps updates the factors by
\begin{equation}
  \begin{bmatrix}\delta\vv\\ \delta\uu\end{bmatrix}
  =\big(s\,T^{\top}HT+\mu\mathbf{I}\big)^{-1}
   T^{\top}\big(A^{\top}r-s\,H(\uu\otimes\vv)\big).
  \label{eq:sup_gn}
\end{equation}
After each step, $\uu$ and $\vv$ are normalized and their norms are folded into $s$. The
ridge $\mu=10^{-2}$ damps the step and is separate from the penalty $\Lambda$. It fixes
the gauge $(\uu,\vv)\mapsto(\kappa\uu,\vv/\kappa)$, which otherwise makes the system
singular. After the initialization and after every step, $s$ is clipped to
$s_{\max}=4$.

\noindent\textbf{Resolution and cost.}
The solve uses bilinear $64\times64$ resizes of $J$, $I$ and the basis maps. After the
resize, $I$ and $b_{\alpha}J$ are floored at $10^{-6}$ before the logarithms of
Eq.~\eqref{eq:residual}. Each training iteration needs one $16\times16$ and four
$8\times8$ linear solves. The veil of Eq.~\eqref{eq:bilinear} is then composed from the
full-resolution basis.

\section{Training Through the Solve}
\label{sup:varpro}

\noindent\textbf{Gradient through the solve.}
The solved coefficients depend on the trained parameters through $A$ (the basis and
$\gamma$) and through $r$ (the render $J$ and the biases $b_{\alpha},\gamma$). We
differentiate each linear solve $x=H^{-1}g$ implicitly. Its backward pass is one more
solve with $H^{\top}$, so we never differentiate through the QR factorization. The power
and Gauss--Newton iterations are unrolled. The damping $\Lambda$ is computed from $A$
inside the graph, so gradients also pass through it.

\noindent\textbf{Variable projection.}
Stage 1 leaves the residual $r^{\star}=R_{\Lambda}r$ of Eq.~\eqref{eq:projector}, with
\begin{equation}
  R_{\Lambda}=\mathbf{I}-A\big(A^{\top}A+\Lambda\big)^{-1}A^{\top}.
  \label{eq:sup_resop}
\end{equation}
Without damping, and for $A$ of full column rank, $R_{0}$ is an orthogonal projector,
and $r^{\star}$ is exactly the part of $r$ that the basis cannot explain. Solving for
the linear coefficients in closed form and optimizing the other parameters on what
remains is variable projection~\cite{Golub1973VarPro}. In this picture, a detached solve
corresponds to alternating minimization, and a learned coefficient table to joint
descent. Ruhe and Wedin~\cite{Ruhe1980Separable} analyse both schemes for separable
least squares. With damping, $R_{\Lambda}$ is no longer a projector. Let $\sigma$ be a
singular value of $A$ after its columns are scaled to unit norm, and $w$ its left
singular vector. Neglecting $\epsilon$, $R_{\Lambda}w=\frac{\lambda}{\sigma^{2}+\lambda}\,w$.
Damping thus shrinks what the basis can explain instead of removing it. The shrinking is
strong along directions that the basis represents well ($\sigma^{2}\gg\lambda$) and weak
along the others.

\noindent\textbf{What reaches the scene.}
Take a parameter that enters only through $r$, such as the Gaussians, and a loss
$\ell(r^{\star})$ with gradient $\nabla\ell$. A detached solve passes $\nabla\ell$ to $r$
unchanged. Training through the solve passes
\begin{equation}
  \frac{\partial\ell}{\partial r}=R_{\Lambda}^{\top}\nabla\ell=R_{\Lambda}\nabla\ell.
  \label{eq:sup_varpro}
\end{equation}
The solve thus filters the gradient with the same matrix that filters the residual.
Without damping and the rank constraint, training through the fit gives the Gaussians,
to first order, only the part of $r$ that the veil cannot explain. With our damping
($\lambda=4$), the part that the veil can explain is shrunk rather than removed. In the
same setting, the basis network, which enters through $A$, learns to cover the part of
the observation that the scene does not explain.

\noindent\textbf{Scope of this analysis.}
This view motivates training through the solve, but it does not fully explain its
benefit. If the training loss were the objective of the fit and the fit were exact, the
envelope theorem would give the same gradient for a detached solve. For example, without
damping, $\ell=\frac{1}{2}\|r^{\star}\|^{2}$ has $\nabla\ell=r^{\star}$, which is already
orthogonal to the range of $A$, so $R_{0}\nabla\ell=\nabla\ell$. In VeilGS, the two
gradients differ for two reasons. First, $\mathcal{L}_{\mathrm{rec}}$ is not the
objective of the fit. It compares intensities at full resolution with an $\ell_1$ and an
SSIM term, after the tone curve of Eq.~\eqref{eq:sup_blend} and clipping. The fit instead
solves a damped least-squares problem in the log domain at $64\times64$. Second, the fit
is not exact. The rank-one stages are nonlinear and stop after four steps. Even at
convergence, their residual would be orthogonal only to the seven tangent directions of
the rank-one model, not to all 16 columns of $A$. We therefore show the benefit of
training through the solve empirically, in Table~\ref{tab:abl_mech}.

\noindent\textbf{The role of damping.}
Damping sets how much of the residual the veil may claim. With $\Lambda=0$, the veil
claims every residual in the range of its basis. Early in training, $J$ is still far
from the scene, and this range covers most of the observation, so the scene receives
little gradient. A large damping pulls the coefficients toward zero. The veil then
reduces to its bias, and the spatial variation of the light moves into the scene.
Table~\ref{tab:sup_damping} and Fig.~\ref{fig:sup_damping} show both effects on the four
scenes of the component ablation. Removing the damping costs 4.82\,dB PSNR and 17.7
MUSIQ, as much as removing the spatial variation of the veil. The veil then absorbs the
scene instead of the light: on \textsc{kitchen} the render loses its structure, and on
\textsc{tree} training diverges to a black render. Doubling the damping to $\lambda=8$
costs 0.62\,dB. Our $\lambda=4$ is best on all six metrics.

\begin{table}[t]
\centering\scriptsize
\setlength{\tabcolsep}{2pt}
\caption{\textbf{Damping of the inner solve}, over the four scenes of
Table~\ref{tab:abl_comp}. The first row is the \emph{w/o damping} row of that table.
Its NIQE and PI average over three scenes, since the undamped \textsc{tree} render is
black. Red: best; yellow: second best.}
\label{tab:sup_damping}
\resizebox{\columnwidth}{!}{\begin{tabular}{l|ccc|ccc}
\hline
 & \multicolumn{3}{c|}{Reconstruction} & \multicolumn{3}{c}{No reference} \\
Variant & PSNR$\uparrow$ & SSIM$\uparrow$ & LPIPS$\downarrow$ & NIQE$\downarrow$ & MUSIQ$\uparrow$ & PI$\downarrow$ \\
\hline
$\lambda = 0$ (w/o damping) & 21.91 & 0.654 & 0.336 & 3.71 & 52.53 & 3.79 \\
$\lambda = 8$ & \ranksecond{26.11} & \ranksecond{0.824} & \ranksecond{0.109} & \ranksecond{2.63} & \ranksecond{69.23} & \ranksecond{2.79} \\
\hline
$\lambda = 4$ (ours) & \rankfirst{26.73} & \rankfirst{0.836} & \rankfirst{0.107} & \rankfirst{2.56} & \rankfirst{70.25} & \rankfirst{2.76} \\
\hline
\end{tabular}
}
\end{table}

\begin{figure}[t]
\centering
\setlength{\tabcolsep}{1pt}
\renewcommand{\arraystretch}{0.5}
\begin{tabular}{cccc}
\includegraphics[width=0.24\columnwidth]{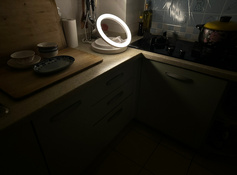} & \includegraphics[width=0.24\columnwidth]{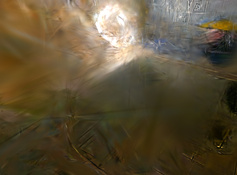} &
\includegraphics[width=0.24\columnwidth]{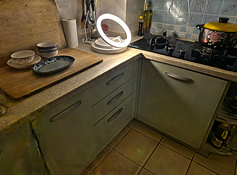} & \includegraphics[width=0.24\columnwidth]{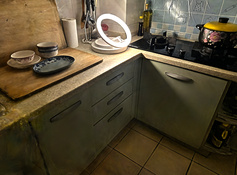} \\
\includegraphics[width=0.24\columnwidth]{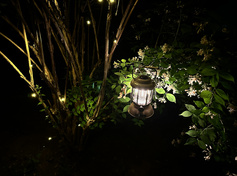} & \includegraphics[width=0.24\columnwidth]{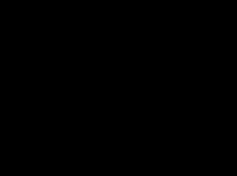} &
\includegraphics[width=0.24\columnwidth]{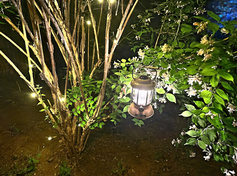} & \includegraphics[width=0.24\columnwidth]{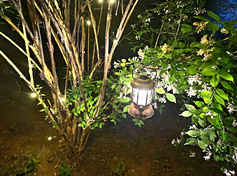} \\
\small Dark input & \small $\lambda=0$ &
\small $\lambda=4$ (ours) & \small $\lambda=8$ \\
\end{tabular}
\caption{\textbf{Damping of the inner solve.} Restored renders of a held-out view of
\textsc{kitchen} (top) and \textsc{tree} (bottom). Without damping, the veil absorbs the
scene: \textsc{kitchen} loses its structure, and \textsc{tree} collapses to a black
render.}
\label{fig:sup_damping}
\end{figure}

\section{Implementation Details}
\label{sup:impl}

\noindent\textbf{Training schedule.}
The physical biases start at $b_{\alpha}=0.3$ and $\gamma=2.2$. For the first 200
iterations, only the biases are trained. The basis network stays fixed until iteration
5{,}000, so that the global level and the geometry settle before the veil gains spatial
structure. The veil is optimized with Adam. Its learning rate decays from $10^{-3}$ to
$10^{-5}$, and its gradients are clipped to norm 1.

\noindent\textbf{Keeping the biases valid.}
Gradients through the solve also reach $b_{\alpha}$ and $\gamma$, and these updates can
be large. The forward model clips $\alpha$ and $\gamma$, and a clip passes no gradient
outside its range, so a bias that crosses its bound never returns. For example, if
$b_{\alpha}$ becomes negative, $\alpha$ sits at its floor everywhere, and neither the bias
nor the basis network receives gradient. We observed this on one night scene within the
first 5{,}000 iterations. After every optimizer step, we therefore project the biases
back into their ranges,
\begin{equation}
  b_{\alpha}\leftarrow\max(b_{\alpha},10^{-3}),\qquad
  \gamma\leftarrow\operatorname{clip}(\gamma,\,1+10^{-3},\,4-10^{-3}).
  \label{eq:sup_projection}
\end{equation}
This turns the clip into projected gradient descent and keeps the clips of the forward
model inactive.

\noindent\textbf{Albedo and shading.}
The albedo $a_i$ is a free parameter, initialized to the colour of its SfM point. The
colour $a_i\odot s_i$ of each Gaussian is clipped at zero before rasterization, so
$J\ge0$. The range term of Eq.~\eqref{eq:albedo} keeps the albedo close to $[0,1]$,
with $\rho_{[l,h]}(x)=\operatorname{relu}(l-x)+\operatorname{relu}(x-h)$ applied per
channel. The shading prior of Eq.~\eqref{eq:shading} is evaluated with the viewing
directions of the current training view, so over training it covers all training views.
In Eqs.~\eqref{eq:albedo} and~\eqref{eq:shading}, the range terms and the squared
norm $\|s_i-\bar{s}_i\mathbf{1}\|^2$ are averaged over the three channels, and the
neighbour sum over the $k=8$ neighbours. The weights $\lambda_{\mathrm{alb}}$,
$\lambda_{\mathrm{sh}}$, $\mu_a$ and $\mu_s$ refer to this normalization. The albedo
smoothness uses the $k=8$ nearest Gaussians in 3D, with weights
$w_{ii'}=\exp(-d_{ii'}/0.1)$ for their distance $d_{ii'}$. The smoothness term of
Eq.~\eqref{eq:albedo} and both terms of Eq.~\eqref{eq:shading} use 500 randomly
chosen Gaussians.

\noindent\textbf{Depth.}
Following~\cite{Kerbl2024Hierarchical}, we align each monocular depth map to the SfM
points with a scale and an offset. A view is skipped if its scale is more than five
times larger or smaller than the median scale. We then min-max normalize the rendered and the
monocular inverse depths to $[0,1]$ in each view. Eq.~\eqref{eq:depth} averages their
absolute difference over all pixels.

\noindent\textbf{Efficiency.}
Table~\ref{tab:sup_efficiency} compares the training time and rendering speed of the
methods of Fig.~\ref{fig:teaser} on LOM. All of them were run on one NVIDIA H100 at
the LOM resolution of $500\times375$.

\begin{table}[t]
\centering\scriptsize
\setlength{\tabcolsep}{4pt}
\caption{\textbf{Efficiency on LOM}: mean training time per scene and rendering speed,
on one NVIDIA H100 at $500\times375$. Red: best; yellow: second best.}
\label{tab:sup_efficiency}
\begin{tabular}{l|cc}
\hline
Method & Training (min)$\downarrow$ & FPS$\uparrow$ \\
\hline
Luminance-GS~\cite{Cui2025LuminanceGS} & \rankfirst{5.1} & \ranksecond{397} \\
LITA-GS~\cite{Zhou2025LITAGS} & \ranksecond{18.6} & 72 \\
I\textsuperscript{2}-NeRF~\cite{Liu2025I2NeRF} & 32.1 & 0.57 \\
\hline
VeilGS (ours) & 20.7 & \rankfirst{407} \\
\hline
\end{tabular}

\end{table}

\section{Evaluation Details and Additional Results}
\label{sup:results}

\noindent\textbf{Scoring.}
On both benchmarks, we average each metric over the test views of a scene and then over
the scenes. On Gaussian-DK, PSNR, SSIM and LPIPS (AlexNet) use the metric code released
with Gaussian-DK~\cite{Ye2024GaussianDK}. NIQE, MUSIQ, CLIP-IQA and PI use the pyiqa
toolbox (v0.1.13). We score the restored render of every method with the same code.

\noindent\textbf{Baselines.}
We train 3DGS~\cite{Kerbl2023}, Gaussian-DK~\cite{Ye2024GaussianDK},
LITA-GS~\cite{Zhou2025LITAGS} and Luminance-GS~\cite{Cui2025LuminanceGS} with their
official code on the standard Gaussian-DK split. The restored image is the light-up
image for Gaussian-DK, the enhanced render for LITA-GS and Luminance-GS, and the render
for 3DGS. For VeilGS, it is $J$, rendered from the camera pose.

\noindent\textbf{Target-conditioned reconstruction.}
Gaussian-DK and RestorGS also evaluate how well a method reproduces the held-out dark
views. Table~\ref{tab:sup_dk_recon} gives this comparison per scene. To render a dark
view, VeilGS keeps all trained parameters fixed and solves the veil from the dark image
of that view, as in training. Gaussian-DK uses the exposure settings of the view, and
the other methods we trained render it from the camera pose. To compare all methods
fairly, we therefore rely on no-reference metrics and the user study on Gaussian-DK
(Table~\ref{tab:dk_noref} of the main paper). These score the restored renders, and
the restored render of VeilGS, $J$, comes from the camera pose alone. In
Table~\ref{tab:sup_dk_recon}, VeilGS has the highest PSNR on every scene and the best
mean PSNR, SSIM and LPIPS.

\noindent\textbf{More qualitative results.}
Fig.~\ref{fig:sup_qual_dk} shows four more Gaussian-DK scenes, and
Fig.~\ref{fig:sup_qual_lom} the two LOM scenes not shown in the main paper.

\begin{table*}[t]
\centering\scriptsize
\setlength{\tabcolsep}{2.2pt}
\caption{\textbf{Target-conditioned reconstruction on Gaussian-DK}, per scene: each
held-out dark view against its dark rendering. All methods use the standard test views
of Gaussian-DK. $\dagger$: results reported in~\cite{Ye2024GaussianDK} (NeRF-W)
and~\cite{Qiao2025RestorGS} (RestorGS). Neither paper reports \textsc{snowman}, so the
table covers the other eleven scenes; the last row averages all twelve, \textsc{snowman}
included, for the methods we ran. Red: best; yellow: second best.}
\label{tab:sup_dk_recon}
\resizebox{\textwidth}{!}{\begin{tabular}{l|ccc|ccc|ccc|ccc|ccc|ccc|ccc}
\hline
Scene & \multicolumn{3}{c|}{3DGS~\cite{Kerbl2023}} & \multicolumn{3}{c|}{Luminance-GS~\cite{Cui2025LuminanceGS}} & \multicolumn{3}{c|}{LITA-GS~\cite{Zhou2025LITAGS}} & \multicolumn{3}{c|}{NeRF-W~\cite{MartinBrualla2021NeRFW}$^{\dagger}$} & \multicolumn{3}{c|}{Gaussian-DK~\cite{Ye2024GaussianDK}} & \multicolumn{3}{c|}{RestorGS~\cite{Qiao2025RestorGS}$^{\dagger}$} & \multicolumn{3}{c}{VeilGS (ours)} \\
 & PSNR$\uparrow$ & SSIM$\uparrow$ & LPIPS$\downarrow$ & PSNR$\uparrow$ & SSIM$\uparrow$ & LPIPS$\downarrow$ & PSNR$\uparrow$ & SSIM$\uparrow$ & LPIPS$\downarrow$ & PSNR$\uparrow$ & SSIM$\uparrow$ & LPIPS$\downarrow$ & PSNR$\uparrow$ & SSIM$\uparrow$ & LPIPS$\downarrow$ & PSNR$\uparrow$ & SSIM$\uparrow$ & LPIPS$\downarrow$ & PSNR$\uparrow$ & SSIM$\uparrow$ & LPIPS$\downarrow$ \\
\hline
\textsc{alley} & 17.19 & 0.637 & 0.191 & 16.09 & 0.612 & 0.226 & 17.04 & 0.617 & 0.210 & 21.59 & 0.646 & 0.331 & 21.72 & 0.774 & 0.141 & \ranksecond{22.18} & \ranksecond{0.781} & \ranksecond{0.138} & \rankfirst{23.75} & \rankfirst{0.807} & \rankfirst{0.106} \\
\textsc{bicycle} & 18.69 & 0.601 & 0.226 & 19.91 & 0.598 & 0.222 & 20.64 & 0.617 & 0.189 & 23.56 & 0.717 & 0.238 & 24.90 & 0.806 & 0.129 & \ranksecond{24.93} & \ranksecond{0.811} & \ranksecond{0.127} & \rankfirst{26.52} & \rankfirst{0.860} & \rankfirst{0.114} \\
\textsc{dormitory} & 19.71 & 0.655 & 0.164 & 21.38 & 0.672 & 0.179 & 21.22 & 0.666 & 0.158 & 24.51 & 0.820 & 0.149 & 25.37 & 0.854 & \ranksecond{0.088} & \ranksecond{25.95} & \ranksecond{0.859} & \ranksecond{0.088} & \rankfirst{27.63} & \rankfirst{0.891} & \rankfirst{0.075} \\
\textsc{fence} & 19.01 & 0.583 & 0.225 & 20.26 & 0.609 & 0.227 & 20.67 & 0.631 & 0.229 & 23.97 & 0.724 & 0.214 & \ranksecond{25.13} & 0.801 & 0.129 & 25.05 & \ranksecond{0.803} & \ranksecond{0.128} & \rankfirst{26.38} & \rankfirst{0.816} & \rankfirst{0.116} \\
\textsc{flower} & 17.68 & 0.586 & 0.182 & 17.89 & 0.577 & 0.195 & 19.34 & 0.622 & 0.187 & 21.66 & 0.662 & 0.223 & 21.70 & 0.735 & 0.133 & \ranksecond{21.97} & \ranksecond{0.742} & \ranksecond{0.127} & \rankfirst{22.95} & \rankfirst{0.760} & \rankfirst{0.117} \\
\textsc{kitchen} & 18.60 & 0.691 & 0.183 & 17.77 & 0.702 & 0.220 & 19.48 & 0.730 & 0.179 & 24.84 & 0.812 & 0.205 & 25.04 & 0.839 & 0.118 & \ranksecond{25.86} & \ranksecond{0.854} & \ranksecond{0.099} & \rankfirst{28.60} & \rankfirst{0.872} & \rankfirst{0.094} \\
\textsc{livingroom} & 20.52 & 0.734 & 0.159 & 19.35 & 0.706 & 0.196 & 21.26 & 0.742 & 0.189 & 22.43 & 0.749 & 0.243 & 24.53 & 0.825 & 0.126 & \ranksecond{24.73} & \ranksecond{0.830} & \rankfirst{0.088} & \rankfirst{26.10} & \rankfirst{0.842} & \ranksecond{0.103} \\
\textsc{piano} & 19.25 & 0.591 & 0.209 & 20.08 & 0.601 & 0.232 & 21.00 & 0.617 & 0.214 & 25.44 & 0.803 & 0.157 & 27.13 & 0.830 & 0.106 & \ranksecond{27.37} & \ranksecond{0.837} & \ranksecond{0.103} & \rankfirst{29.97} & \rankfirst{0.889} & \rankfirst{0.079} \\
\textsc{staircase} & 16.16 & 0.480 & 0.279 & 18.91 & 0.536 & 0.189 & 20.51 & 0.583 & 0.176 & 23.96 & 0.809 & 0.115 & 26.73 & \ranksecond{0.887} & \rankfirst{0.067} & \ranksecond{27.02} & 0.882 & \rankfirst{0.067} & \rankfirst{27.41} & \rankfirst{0.908} & \ranksecond{0.073} \\
\textsc{street} & 16.74 & 0.539 & 0.266 & 17.32 & 0.553 & 0.296 & 21.51 & 0.661 & 0.221 & 22.85 & 0.638 & 0.293 & 26.06 & \ranksecond{0.786} & 0.144 & \ranksecond{26.44} & \rankfirst{0.793} & \ranksecond{0.141} & \rankfirst{26.45} & 0.782 & \rankfirst{0.129} \\
\textsc{tree} & 18.78 & 0.573 & 0.210 & 18.69 & 0.552 & 0.231 & 19.55 & 0.559 & 0.223 & 21.66 & 0.649 & 0.260 & 23.60 & 0.778 & 0.132 & \ranksecond{23.89} & \ranksecond{0.793} & \rankfirst{0.123} & \rankfirst{24.27} & \rankfirst{0.799} & \ranksecond{0.131} \\
\hline
Mean (11 scenes) & 18.39 & 0.606 & 0.209 & 18.88 & 0.611 & 0.219 & 20.20 & 0.641 & 0.198 & 23.32 & 0.730 & 0.221 & 24.72 & 0.810 & 0.119 & \ranksecond{25.04} & \ranksecond{0.817} & \ranksecond{0.112} & \rankfirst{26.36} & \rankfirst{0.839} & \rankfirst{0.103} \\
Mean (12 scenes) & 18.56 & 0.606 & 0.208 & 19.13 & 0.614 & 0.217 & 20.21 & 0.637 & 0.197 & -- & -- & -- & \ranksecond{24.88} & \ranksecond{0.813} & \ranksecond{0.118} & -- & -- & -- & \rankfirst{26.39} & \rankfirst{0.841} & \rankfirst{0.104} \\
\hline
\end{tabular}
}
\end{table*}

\begin{figure*}[t]
\centering
\setlength{\tabcolsep}{1pt}
\renewcommand{\arraystretch}{0.5}
\begin{tabular}{cccccc}
\includegraphics[width=0.162\textwidth]{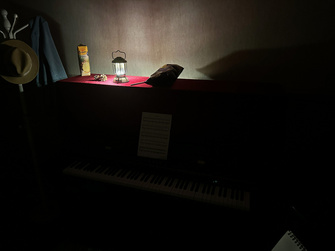} & \includegraphics[width=0.162\textwidth]{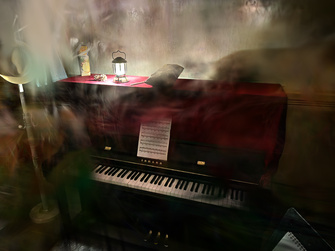} & \includegraphics[width=0.162\textwidth]{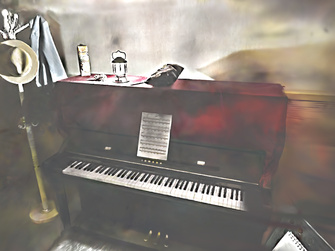} & \includegraphics[width=0.162\textwidth]{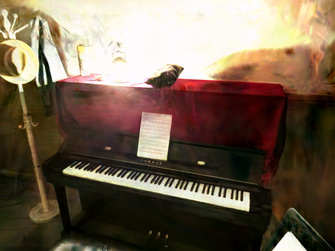} & \includegraphics[width=0.162\textwidth]{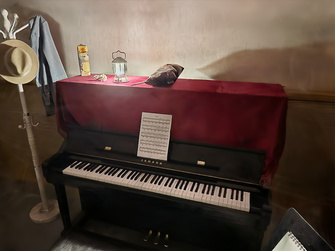} & \includegraphics[width=0.162\textwidth]{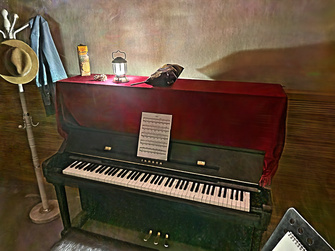} \\
\includegraphics[width=0.162\textwidth]{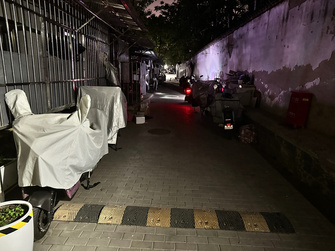} & \includegraphics[width=0.162\textwidth]{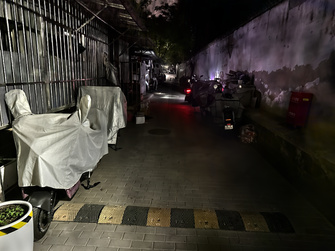} & \includegraphics[width=0.162\textwidth]{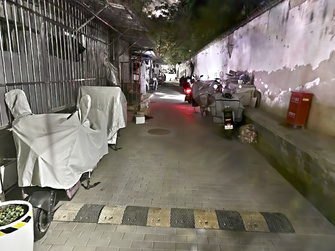} & \includegraphics[width=0.162\textwidth]{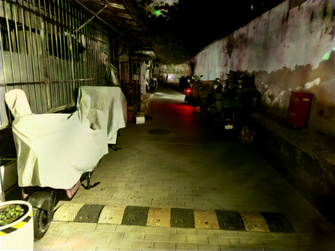} & \includegraphics[width=0.162\textwidth]{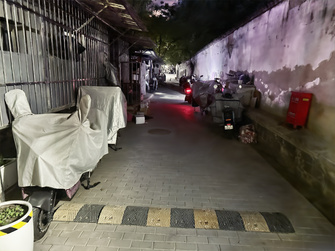} & \includegraphics[width=0.162\textwidth]{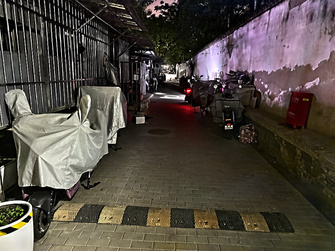} \\
\includegraphics[width=0.162\textwidth]{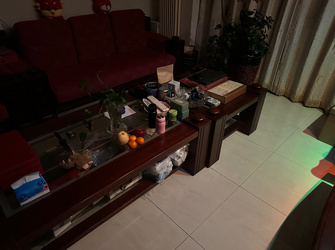} & \includegraphics[width=0.162\textwidth]{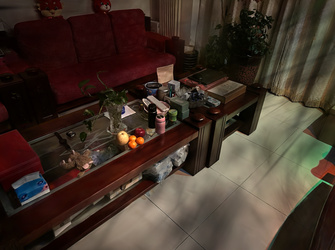} & \includegraphics[width=0.162\textwidth]{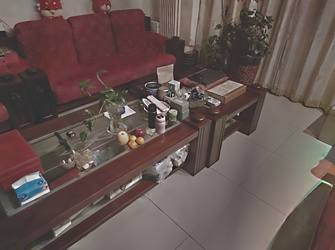} & \includegraphics[width=0.162\textwidth]{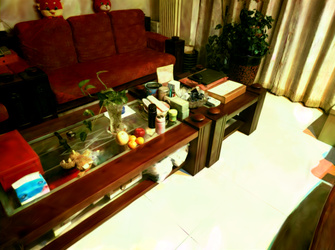} & \includegraphics[width=0.162\textwidth]{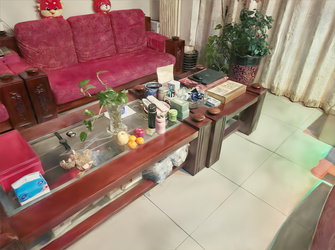} & \includegraphics[width=0.162\textwidth]{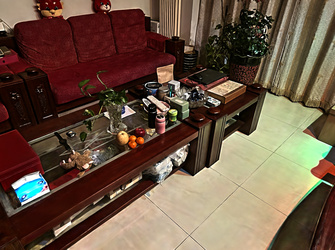} \\
\includegraphics[width=0.162\textwidth]{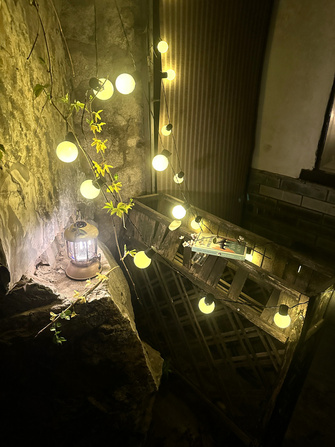} & \includegraphics[width=0.162\textwidth]{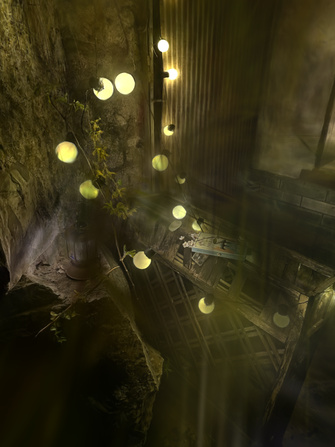} & \includegraphics[width=0.162\textwidth]{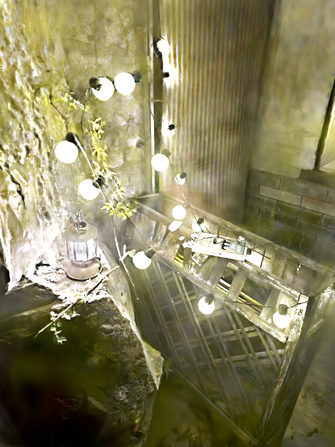} & \includegraphics[width=0.162\textwidth]{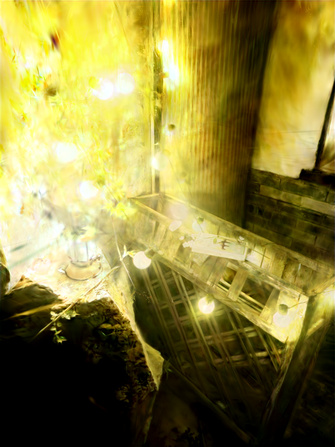} & \includegraphics[width=0.162\textwidth]{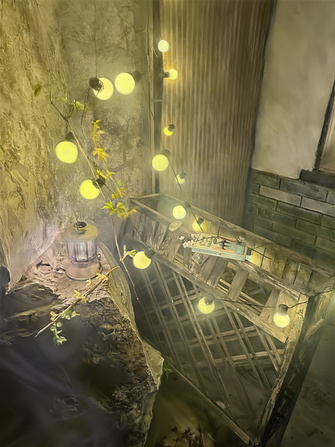} & \includegraphics[width=0.162\textwidth]{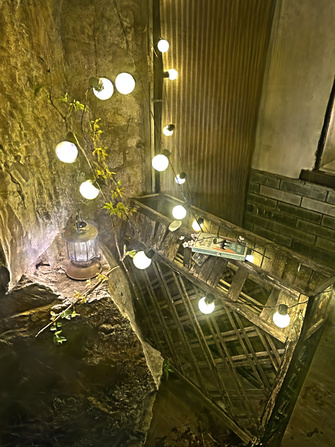} \\
\small Dark input & \small 3DGS~\cite{Kerbl2023} & \small Luminance-GS~\cite{Cui2025LuminanceGS} &
\small LITA-GS~\cite{Zhou2025LITAGS} & \small Gaussian-DK~\cite{Ye2024GaussianDK} & \small VeilGS (ours) \\
\end{tabular}
\caption{\textbf{More qualitative results on Gaussian-DK.} Restored renders of held-out
views of \textsc{piano}, \textsc{alley}, \textsc{livingroom} and \textsc{fence} (top to
bottom). Among the scenes not shown in Figs.~\ref{fig:teaser} and~\ref{fig:qual_dk},
these four have the largest PSNR lead of VeilGS in Table~\ref{tab:sup_dk_recon}. Each row
shows the held-out view with the largest lead over the baselines we trained.}
\label{fig:sup_qual_dk}
\end{figure*}

\begin{figure*}[t]
\centering
\setlength{\tabcolsep}{1pt}
\renewcommand{\arraystretch}{0.5}
\begin{tabular}{cccccc}
\includegraphics[width=0.162\textwidth]{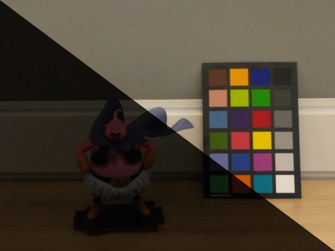} & \includegraphics[width=0.162\textwidth]{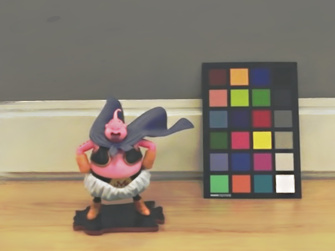} & \includegraphics[width=0.162\textwidth]{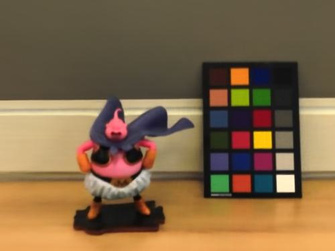} &
\includegraphics[width=0.162\textwidth]{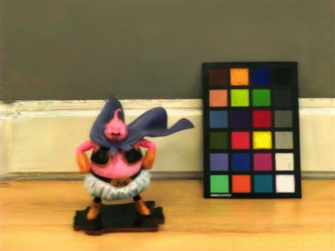} & \includegraphics[width=0.162\textwidth]{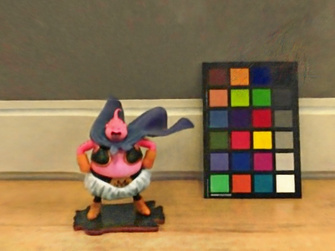} & \includegraphics[width=0.162\textwidth]{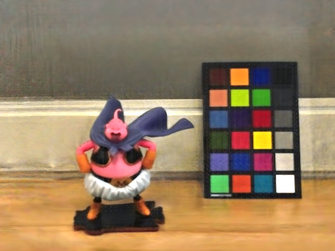} \\
\includegraphics[width=0.162\textwidth]{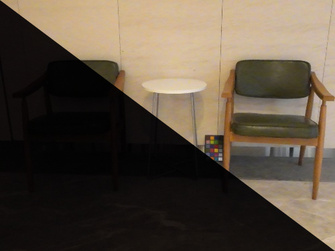} & \includegraphics[width=0.162\textwidth]{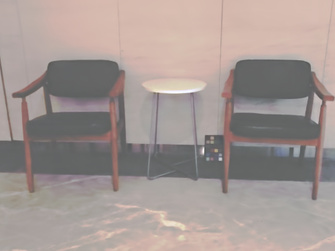} & \includegraphics[width=0.162\textwidth]{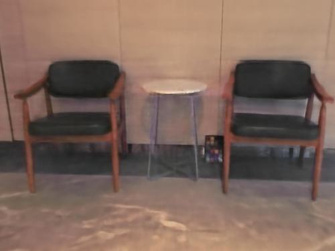} &
\includegraphics[width=0.162\textwidth]{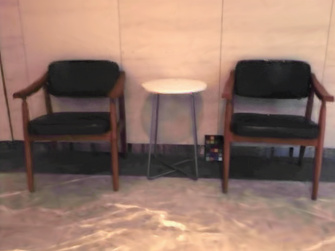} & \includegraphics[width=0.162\textwidth]{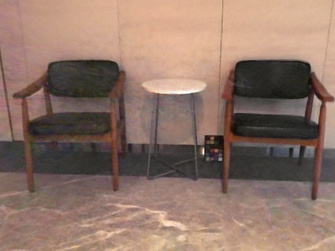} & \includegraphics[width=0.162\textwidth]{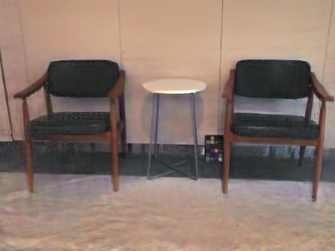} \\
\small Low / Normal light & \small Luminance-GS~\cite{Cui2025LuminanceGS} & \small Aleth-NeRF~\cite{Cui2024AlethNeRF} &
\small LITA-GS~\cite{Zhou2025LITAGS} & \small I\textsuperscript{2}-NeRF~\cite{Liu2025I2NeRF} & \small VeilGS (ours) \\
\end{tabular}
\caption{\textbf{More qualitative results on LOM.} \textsc{buu} (top) and \textsc{chair}
(bottom), the two scenes of Table~\ref{tab:lom} not shown in Fig.~\ref{fig:qual_lom}.
The first column shows the low-light input (lower left) and the normal-light frame
(upper right).}
\label{fig:sup_qual_lom}
\end{figure*}

\section{User Study}
\label{sup:study}

\noindent\textbf{Setup.}
We ran a blind rating study on a web page. Each question shows five versions of one
scene, labelled A to E, and hides the method behind each letter. The assignment of
methods to letters and the order of the scenes are shuffled for every participant. The
first part covers the twelve Gaussian-DK scenes with looping videos of
3DGS~\cite{Kerbl2023}, Luminance-GS~\cite{Cui2025LuminanceGS},
LITA-GS~\cite{Zhou2025LITAGS}, Gaussian-DK~\cite{Ye2024GaussianDK} and VeilGS, played in
sync. In each video, the camera moves back and forth between two held-out views of the
scene (on \textsc{snowman}, a held-out and a training view). The path is the same for
every method, and the videos run at the quarter resolution of the benchmark and 30
frames per second for 8.9\,s. The second part covers the five LOM scenes with still
images of Luminance-GS, LITA-GS, Aleth-NeRF~\cite{Cui2024AlethNeRF},
I\textsuperscript{2}-NeRF~\cite{Liu2025I2NeRF} and VeilGS at $500\times375$. It uses the
views of Figs.~\ref{fig:qual_lom} and~\ref{fig:sup_qual_lom} and one test view of
\textsc{shrub}. The Aleth-NeRF and I\textsuperscript{2}-NeRF images are the renders
released by their authors. The other baselines are our runs of their official code.

\noindent\textbf{Task.}
Participants rate every version for overall visual quality on a five-point scale:
5 Excellent (looks like a clean, well-lit photo or video), 4 Good (minor flaws you have to
look for), 3 Fair (noticeable flaws, but the scene is clear), 2 Poor (obvious flaws that
get in the way) and 1 Bad (hard to make out the scene). The instructions ask whether the
scene is clearly visible, whether its colours and brightness look natural, and whether it
is free of haze, blur, noise, floating blobs or flicker. Participants can enlarge any
version and switch between the versions of a question. Several versions may receive the
same score, and ratings can be changed until submission. Participants are asked to keep
their screen at full brightness.

\noindent\textbf{Consent and ethics.}
Before starting, participants agree that their anonymous ratings may be used for
research. The study recorded no identifying information. No ethics review was carried
out, as we judged the study to be of low risk.

\noindent\textbf{Analysis.}
In total, 55 participants finished the study and rated every version. This gives
4{,}675 ratings, given in a median of 8.4 minutes (interquartile range 5.8--14.2). If a
version was rated more than once, the last rating counts. The ratings of one participant
are correlated, so we compute every statistic over participants. The mean opinion score
(MOS) of a method is the mean, over participants, of the average score that each
participant gave it. Its 95\% confidence interval is $t\,\sigma/\sqrt{n}$ over the $n=55$
participant averages. We compare VeilGS with each baseline by a two-sided Wilcoxon
signed-rank test on the same participant averages.

\begin{figure*}[t]
\centering
\includegraphics[width=\textwidth]{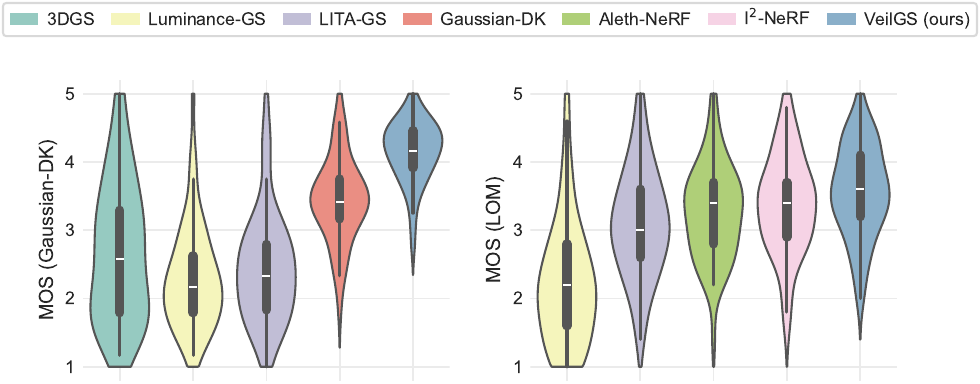}
\caption{\textbf{Scores of the user study.} Each violin shows, for one method, the
average score that each of the 55 participants gave it over the scenes of a benchmark.
The box marks the quartiles and the white line the median. The means and their 95\%
confidence intervals are in Tables~\ref{tab:sup_mos_dk} and~\ref{tab:sup_mos_lom}.}
\label{fig:sup_mos}
\end{figure*}

\noindent\textbf{Results.}
Tables~\ref{tab:sup_mos_dk} and~\ref{tab:sup_mos_lom} give the MOS on each benchmark,
and Table~\ref{tab:dk_noref} of the main paper the MOS per scene on Gaussian-DK.
Fig.~\ref{fig:sup_mos} shows how the scores of the participants are distributed. On
the twelve Gaussian-DK scenes, VeilGS has a MOS of 4.15, against
3.44 for Gaussian-DK. Of the 55 participants, 51 rate it higher than Gaussian-DK
($p<10^{-9}$). It has the highest score on ten of the twelve scenes; on \textsc{piano}
and \textsc{snowman}, Gaussian-DK scores higher. On LOM, VeilGS has the highest MOS,
3.57, against 3.38 for I\textsuperscript{2}-NeRF ($p=0.001$) and 3.27 for Aleth-NeRF. It
is best on two of the five scenes, Aleth-NeRF on \textsc{buu} and \textsc{sofa}, and
I\textsuperscript{2}-NeRF on \textsc{shrub}. On both benchmarks, the per-participant
difference favours VeilGS against every baseline ($p\le0.002$).

\begin{table}[t]
\centering\scriptsize
\setlength{\tabcolsep}{4pt}
\caption{\textbf{Mean opinion score on Gaussian-DK} (1--5), with the 95\% confidence
interval over participants. Videos of all twelve scenes; Table~\ref{tab:dk_noref}
gives the scores per scene. Red: best; yellow: second best.}
\label{tab:sup_mos_dk}
\begin{tabular}{l|c}
\hline
Method & MOS $\pm$ 95\% CI \\
\hline
3DGS~\cite{Kerbl2023} & 2.68 $\pm$ 0.27 \\
Luminance-GS~\cite{Cui2025LuminanceGS} & 2.31 $\pm$ 0.20 \\
LITA-GS~\cite{Zhou2025LITAGS} & 2.45 $\pm$ 0.22 \\
Gaussian-DK~\cite{Ye2024GaussianDK} & \ranksecond{3.44} $\pm$ 0.17 \\
\hline
VeilGS (ours) & \rankfirst{4.15} $\pm$ 0.12 \\
\hline
\end{tabular}

\end{table}

\begin{table}[t]
\centering\scriptsize
\setlength{\tabcolsep}{3pt}
\caption{\textbf{Mean opinion score on LOM} (1--5), per scene and on average, with the
95\% confidence interval of the average over participants. Red: best; yellow: second
best.}
\label{tab:sup_mos_lom}
\resizebox{\columnwidth}{!}{\begin{tabular}{l|ccccc|c}
\hline
Method & \textsc{bike} & \textsc{buu} & \textsc{chair} & \textsc{shrub} & \textsc{sofa} & Mean $\pm$ 95\% CI \\
\hline
Luminance-GS~\cite{Cui2025LuminanceGS} & 1.82 & 2.76 & 2.24 & 1.91 & 2.42 & 2.23 $\pm$ 0.25 \\
LITA-GS~\cite{Zhou2025LITAGS} & 2.71 & 2.95 & \ranksecond{3.55} & 2.35 & \ranksecond{4.00} & 3.11 $\pm$ 0.22 \\
Aleth-NeRF~\cite{Cui2024AlethNeRF} & 2.22 & \rankfirst{3.64} & 2.95 & \ranksecond{3.42} & \rankfirst{4.13} & 3.27 $\pm$ 0.19 \\
I\textsuperscript{2}-NeRF~\cite{Liu2025I2NeRF} & \ranksecond{3.82} & \ranksecond{3.60} & 2.84 & \rankfirst{4.00} & 2.65 & \ranksecond{3.38} $\pm$ 0.20 \\
\hline
VeilGS (ours) & \rankfirst{4.05} & 3.42 & \rankfirst{3.85} & 2.78 & 3.75 & \rankfirst{3.57} $\pm$ 0.18 \\
\hline
\end{tabular}
}
\end{table}

\clearpage
\end{document}